\documentclass{article}

\usepackage[nonatbib,final]{neurips_2020}

\usepackage[utf8]{inputenc} % allow utf-8 input
\usepackage[T1]{fontenc}    % use 8-bit T1 fonts
\usepackage{hyperref}       % hyperlinks
\usepackage{url}            % simple URL typesetting
\usepackage{booktabs}       % professional-quality tables
\usepackage{amsfonts}       % blackboard math symbols
\usepackage{nicefrac}       % compact symbols for 1/2, etc.
\usepackage{microtype}      % microtypography

\usepackage{xcolor}
\usepackage{amsmath}
\usepackage{breqn}
\usepackage{float}
\usepackage{graphicx}
\usepackage{caption}
\usepackage{subcaption}
\usepackage[cal=cm]{mathalfa}
\usepackage{comment}
\usepackage{mathtools}
\usepackage{bbm}
\usepackage{multirow}
\usepackage{diagbox}

\newcommand{\ours}{DVERGE}

\usepackage[ruled]{algorithm}
\usepackage[noend]{algpseudocode}
\usepackage[normalem]{ulem}
\makeatletter

\DeclareMathOperator*{\argmax}{argmax}

\newcommand{\x}{$\times$}
\newcommand{\printfnsymbol}[1]{%
  \textsuperscript{\@fnsymbol{#1}}%
}
\def\algbackskip{\hskip-\ALG@thistlm}
\makeatother

\graphicspath{ {fig/} }

\title{DVERGE: Diversifying Vulnerabilities for Enhanced Robust Generation of Ensembles}

\author{
Huanrui Yang\textsuperscript{\rm 1}\thanks{Equal contribution.}, Jingyang Zhang\textsuperscript{\rm 1}\footnotemark[1], Hongliang Dong\textsuperscript{\rm 1}\footnotemark[1], Nathan Inkawhich\textsuperscript{\rm 1},\\ \textbf{Andrew Gardner\textsuperscript{\rm 2}, Andrew Touchet\textsuperscript{\rm 2}, Wesley Wilkes\textsuperscript{\rm 2}, Heath Berry\textsuperscript{\rm 2}, Hai Li\textsuperscript{\rm 1}}\\ 
\textsuperscript{\rm 1}Department of Electrical and Computer Engineering, Duke University\\
\textsuperscript{\rm 2}Radiance Technologies\\
\textsuperscript{\rm 1}\{huanrui.yang, jz288, hongliang.dong, nai2, hai.li\}@duke.edu,\\ \textsuperscript{\rm 2}\{andrew.gardner, atouchet, Wesley.Wilkes, Heath.Berry\}@radiancetech.com
}

\begin{document}

\maketitle

\begin{abstract}
Recent research finds CNN models for image classification demonstrate overlapped adversarial vulnerabilities: adversarial attacks can mislead CNN models with small perturbations, which can effectively transfer between different models trained on the same dataset.
%Adversarial training is designed for general robustness improvement, not dedicated for transfer attack. 
Adversarial training, as a general robustness improvement technique, eliminates the vulnerability in a single model by forcing it to learn robust features. 
The process is hard, often requires models with large capacity, and suffers from  significant loss on clean data accuracy.
Alternatively, ensemble methods are proposed to induce sub-models with diverse outputs against a transfer adversarial example, making the ensemble robust against transfer attacks even if each sub-model is individually non-robust. 
Only small clean accuracy drop is observed in the process.
However, previous ensemble training methods are not efficacious in inducing such diversity and thus ineffective on reaching robust ensemble. 
We propose \ours{}, which isolates the adversarial vulnerability in each sub-model by distilling non-robust features, and diversifies the adversarial vulnerability to induce diverse outputs against a transfer attack. The novel diversity metric and training procedure enables \ours{} to achieve higher robustness against transfer attacks comparing to previous ensemble methods, and enables the improved robustness when more sub-models are added to the ensemble. 
The code of this work is available at \url{https://github.com/zjysteven/DVERGE}.
%Higher white-box robustness can also be achieved by amending the proposed method with adversarial training.
\end{abstract}

\section{Introduction}
\label{sec:intro}

\begin{figure}[tb]
\centering
  %\rule{12cm}{3cm}
  \captionsetup{width=.95\linewidth}
  \includegraphics[width=.75\linewidth]{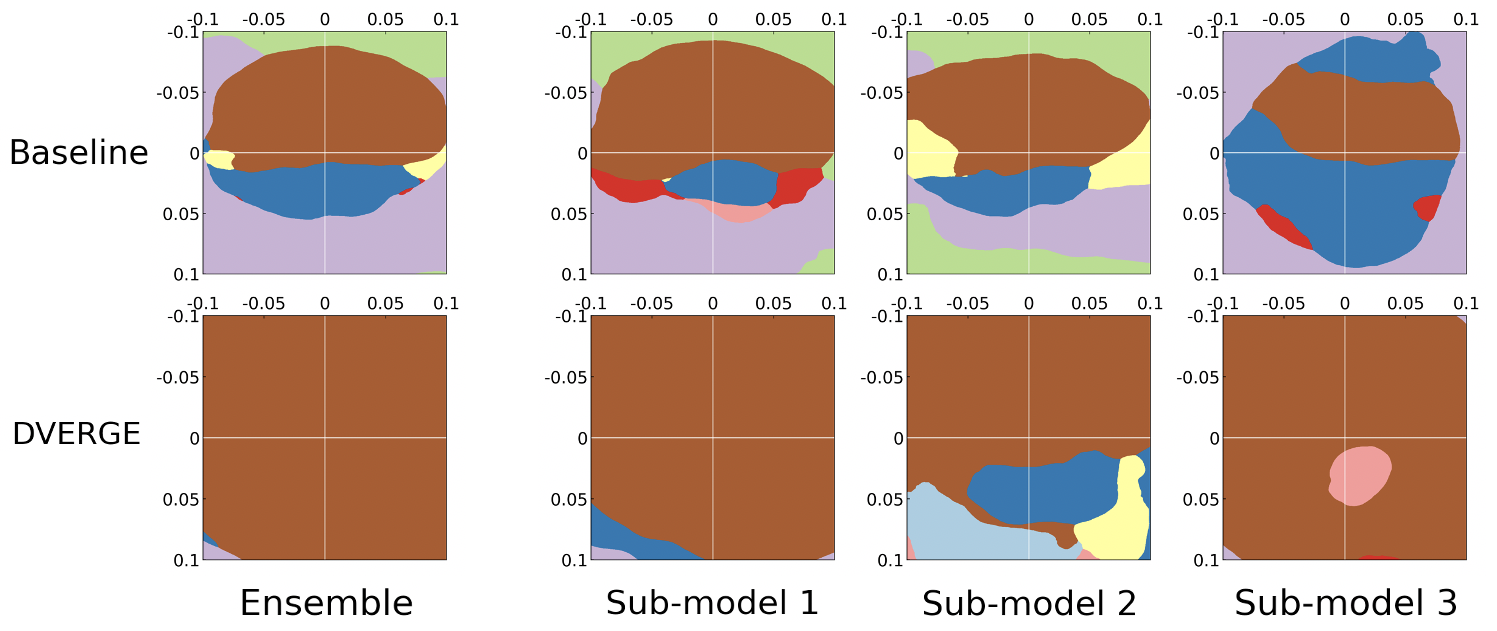}
  \caption{Decision regions in the $\ell_\infty$ ball around the same testing image learned by ensembles of 3 ResNet-20 models trained on CIFAR-10 dataset. Same color indicates the same predicted label. The vertical axis is along the adversarial direction of a surrogate benign ensemble, and the horizontal axis is along a random Rademacher vector. The same axes are used for each subplot. Adversarial vulnerability can be inferred from the closest decision boundary and corresponding class. The baseline ensemble is achieved via standard training on clean data while the bottom ensemble is trained with \ours. More plots of this nature can be seen in \textbf{Appendix \ref{ap:decision_region}}.}
  \label{fig:dec}
  \vspace{-10pt}
\end{figure}

Recent discoveries of \textit{adversarial attacks} cast doubt on the inherent robustness of convolutional neural networks (CNNs)~\cite{goodfellow2014explaining,carlini2017towards,madry2018towards}. 
These attacks, commonly referred to as \textit{adversarial examples}, comprise precisely crafted input perturbations that are often imperceptible to humans yet consistently induce misclassification in CNN models. 
Moreover, previous research has demonstrated widespread \textit{transferability} of adversarial examples, wherein adversarial examples generated against an arbitrary model can reliably mislead other unspecified deep learning models trained with the same dataset~\cite{papernot2016transferability,ilyas2019adversarial,Inkawhich2020Transferable}.
% Robust and Non-robust feature
%To better understand the cause and high transferability of adversarial attacks, 
Ilyas et al.~\cite{ilyas2019adversarial} conjecture the existence of robust and non-robust features within standard image classification datasets.
%As human can easily classify the image with ``human-meaningful'' robust features, which are not sensitive to small additive noise, deep learning models tends to learn a set of non-robust features more easily, which informally are features that are highly correlated with the output label but incomprehensible to humans. Adversarial examples therefore tend to exploit such dependency on non-robust features to mislead CNN models~\cite{ilyas2019adversarial}.
Whereas humans may understand an image via ``human-meaningful'' robust features, which usually are insensitive to small additive noise, deep learning models are more prone to learning non-robust features.
Non-robust features are highly correlated with output labels and help improve clean accuracy but are not visually meaningful and are sensitive to noise. 
Such dependency on non-robust features leads to \textit{adversarial vulnerability} that is exploited by adversarial examples to mislead CNN models. %~\cite{ilyas2019adversarial}. 
Moreover, Ilyas et al. empirically show that CNN models independently trained on the same dataset tend to capture similar non-robust features, demonstrating overlapping vulnerability~\cite{ilyas2019adversarial}. 
This property can be observed from the example in the upper row of Figure~\ref{fig:dec}, where an ensemble is trained on clean data and each of its sub-models are vulnerable along the same axis of a transfer attack.
This similarity is key to the high transferability of adversarial attacks~\cite{ilyas2019adversarial,li2015convergent}. 

% Adversarial training: learning robust feature -> difficult
Extensive research has been conducted to improve the robustness of CNN models against adversarial attacks, most notably \textit{adversarial training} \cite{madry2018towards}. 
Adversarial training minimizes the loss of a CNN model on online-generated adversarial examples against itself at each training step. 
%The effectiveness of adversarial training methods is mainly dependent on how the adversarial examples are generated, so multi-step attacks like PGD~\cite{madry2018towards} are often used.
This process forces the model to prefer robust to non-robust features and thereby largely eliminates the model's vulnerability.
Nevertheless, learning robust features is hard, so adversarial training often leads to a significant increase in the generalization error on clean testing data~\cite{tsipras2018robustness}.
%{\RED \textbf{[Need to rework on this part.]} However, for CNN models, learning to classify the inputs with the robust features correctly is harder than capturing the non-robust features of the training set, so the improvement on robustness often comes with a significant drop of the classification accuracy on clean inputs~\cite{tsipras2018robustness}.
%It is relatively easier to capture non-robust features of the training set than to learn robust features that are insensitive to noises. Thus, the improvement on robustness often comes with the classification accuracy drop on clean inputs~\cite{tsipras2018robustness}.
%Previous experiments show that a larger model could better capture robust features and reach higher robustness. %Yet the cost of performing adversarial training with large model is often prohibitive, and 
%Yet the tradeoff between the model capacity and robustness induced by adversarial training is typically inconsistent, e.g., increasing a small number of channels or layers may not lead to visible robustness improvement~\cite{madry2018towards,Xie2020Intriguing}.

% Ensemble: alternative way focus on transferability
%Instead of training a single large model for higher accuracy and robustness, 
Similar to traditional ensemble methods like bagging~\cite{breiman1996bagging} and boosting~\cite{dietterich2000ensemble}, which train an ensemble of weak learners with diverse predictions to improve overall accuracy,
a recent line of research proposes to train an ensemble of individually non-robust sub-models that produce diverse outputs against transferred adversarial examples~\cite{bagnall2017training,pang2019improving,kariyappa2019improving}.
%The approach targets to defend black-box transfer attacks by inducing sub-models to produce different outputs against adversarial examples.
Intuitively, the approach can defend against black-box transfer attacks as an attack can succeed only when multiple sub-models converge towards the same wrong prediction~\cite{kariyappa2019improving}.
%In such case, blackbox transfer attacks can be defended as successful attack incurs only when it fools multiple sub-models towards the same wrong prediction~\cite{kariyappa2019improving}.
Such an ensemble could also hypothetically achieve high clean accuracy since the training process doesn't exclude non-robust features.
%However, the strong transferability of adversarial examples makes it non-trivial to train sub-models to induce diverse outputs.
%Although such ensemble may not be strongly robust to white-box adversarial attacks, it is potentially robust to black-box transfer attacks if the diversity metric can induce different sub-models to have diverse outputs against a transferred adversarial example.
%In such case the attack will only be successful if it can fool multiple sub-models simultaneously~\cite{kariyappa2019improving}.
%However, the strong transferability of adversarial examples makes inducing diverse output against them non-trivial. 
%To limit the transferability of the attacks between sub-models, 
Various ensemble training methods have been explored, such as diversifying output logits’ distributions~\cite{bagnall2017training,pang2019improving} or minimizing
the cosine similarity between the input gradient direction of each sub-model~\cite{kariyappa2019improving}. 
Yet empirical results show that these diversity metrics are not very effective at inducing output diversity among sub-models, and thus the corresponding ensemble can hardly attain the desired robustness~\cite{tramer2020adaptive}.

We note that black-box transfer attacks are prevalent in real-world applications where model parameters are not exposed to end users~\cite{Inkawhich2020Transferable,kariyappa2019improving}. 
Moreover, high clean accuracy is always desirable. %Therefore ensemble training methods that can reach high robustness against transfer attacks while maintaining high clean accuracy are preferred in practice.
% Our method: focus on non-robust features
We therefore seek an effective training method that mitigates attack transferability while maintaining high clean accuracy. %of individually non-robust sub-models. %by closely investigating the cause of adversarial vulnerability in each sub-model. 
%Specifically, 
Based on a close investigation of the cause of adversarial vulnerability in sub-models, we propose to \textit{distill} the features learned by each sub-model corresponding to its vulnerability to adversarial examples and use the overlap between the distilled features to measure the diversity between sub-models.
As adversarial examples exploit the vulnerability of sub-models, a small overlap between sub-models indicates that a successful adversarial example on one sub-model is unlikely to fool the other sub-model. 
Consequently, our method impedes attack transferability between sub-models and leads to diverse outputs against a transferred adversarial example.
%{\RED As adversarial examples exploit the vulnerability of sub-models for successful attacks, a small overlapping suggests that successful adversarial examples on one sub-model will not likely to fool other sub-models, therefore blocking the attack transferability between sub-models and lead to diverse outputs against a transfer adversarial example.}
Based on this diversity metric, we propose \textit{\textbf{D}iversifying \textbf{V}ulnerabilities for \textbf{E}nhanced \textbf{R}obust \textbf{G}eneration of \textbf{E}nsembles} (\ours), which uses a round-robin training procedure to distill and diversify the features corresponding to each sub-model's vulnerability.
%{\RED \textbf{[need to revise]} 
%Since the strong transferability of adversarial example comes with different models capturing similar non-robust features, 
%We propose to measure the overlap between non-robust features learned by different sub-models within an ensemble, and enforce them to utilize different sets of non-robust features.
%Specifically, we propose \textit{Robust Ensemble Generation via Vulnerability Diversification} (\ours), which uses a round-robin non-robust feature diversification procedure to distill and diversify the non-robust feature used by each sub-model.}
The proposed ensemble training method makes the following contributions:
\begin{itemize}
    \item \ours{} can successfully isolate and diversify the vulnerability in each sub-model such that within-ensemble attack transferability is nearly eliminated;
    \item \ours{} can significantly improve the overall robustness of the ensemble against black-box transfer attacks without significantly impacting the clean accuracy;
    \item The diversity induced by \ours{} consistently improves robustness as the number of ensemble sub-models increases under equivalent evaluation conditions.
\end{itemize}

As shown in the bottom row of Figure~\ref{fig:dec}, diverse vulnerabilities allowed to persist in each sub-model for high clean accuracy by \ours{} combine to yield an ensemble robust to transfer attacks.
%Although our diverse ensemble training method mainly addresses black-box transfer attacks, we also show that 
Our method can also be augmented with the adversarial training objective to yield an ensemble with both satisfying white-box robustness and higher clean accuracy compared to exclusively adversarial training.
To the best of our knowledge, this work is the first to utilize distilled features for training diverse ensembles and quantitatively relate it to the robustness against adversarial attacks. 

\section{Related work}
\label{sec:back}
%In this section we review adversarial attacks and concepts of ensemble diversity as a defense against these attacks. 

\paragraph{Adversarial attack and defense.}
The pervasiveness of adversarial examples highlights the vulnerability of modern CNN systems to malicious inputs. 
An adversarial attack usually applies an additive perturbation $\delta$ subject to some constraint $S$ to an original input $x$ to form the adversarial example $x_{adv} = x+\delta$. 
The goal of the attack is to find $\delta$ so that $x_{adv}$ can maximize the loss $\mathcal{L}_\theta$ of some CNN model with parameters $\theta$ w.r.t. $x$'s true label $y$.
The attacker's objective can be formulated as $x_{adv} = x+\argmax_{\delta\in S} \mathcal{L}_\theta(x+\delta, y).$
The constraint $S$ typically ensures adversarial examples are visually indistinguishable from original inputs, which is often defined as $||\delta||_p\leq \epsilon$ for some perturbation strength $\epsilon$ and $\ell_p$-norm, \textit{e.g.}\@ $p=0$~\cite{papernot2016limitations}, $p=2$~\cite{carlini2017towards}, or $p=\infty$~\cite{goodfellow2014explaining,madry2018towards}. 
In this work, we focus on the attack bounded by the $\ell_\infty$ norm, which has become increasingly common in recent attack and defense research studies.
Madry et al.~\cite{madry2018towards} show that the attacker's objective can be effectively optimized in a multi-step projected gradient descent (PGD) manner, where in each step of the gradient update the achieved adversarial example is projected back into the constraint set $S$ to make sure it complies with the $\ell_\infty$ norm constraint. 
The attack can be further strengthened via application of a random starting point~\cite{madry2018towards} or consideration of the gradient's momentum information during the optimization~\cite{dong2017discovering,zheng2019distributionally}.
% Note: do we need to mention other attacks beyond additive noise here? Like patch or functional attack. Personally I don't think so.

Various empirical methods have been investigated for improving model robustness.
%{\RED Some early works detect and restore adversarial examples with pre-processing detectors or denoisers~\cite{denoiser,feinman2017detecting,song2018pixeldefend} or flatten local gradients with defensive distillation~\cite{papernot2016distillation}. 
%These methods, however, are not adequate to defend adaptive attacks~\cite{carlini2017towards,carlini2017adversarial,tramer2020adaptive}.}
Among theses methods, adversarial training~\cite{madry2018towards} has gained prominence for its reliability and effectiveness.
Adversarial training generates adversarial examples while concurrently training CNN model(s) to minimize the loss on these adversarial examples. 
The objective of adversarial training is formulated as a min-max optimization: $\min_{\theta} \mathbbm{E}_{(x,y)\sim D}[\max_{\delta\in S} \mathcal{L}_\theta(x+\delta, y)],$
where the inner maximization is often conducted with PGD attacks for greater robustness~\cite{madry2018towards}. 
Although recent research shows that PGD adversarial training encourages a model to capture robust features within datasets~\cite{ilyas2019adversarial}, the process is difficult and costly. 
The learning of robust feature detrimentally and significantly affects the accuracy of the model on clean data~\cite{tsipras2018robustness}, and the model architecture needs to be much larger in order to compensate for the added complexity of the objective~\cite{madry2018towards}.

\paragraph{Ensemble for improved accuracy and uncertainty measurement.}
Traditionally, ensemble learning methods have been extensively studied to improve the performance of the model and tackle out-of-distribution uncertainty and generalization. With some early success on performance improvement with neural network ensemble~\cite{hansen1990neural}, bagging~\cite{breiman1996bagging} and boosting~\cite{dietterich2000ensemble}, Kuncheva~et~al.~\cite{kuncheva2003measures} make a thorough evaluation on the relationship between sub-model diversity and ensemble accuracy, and find that as a higher diversity generally improves ensemble accuracy, specially designed diversity metric and training algorithm are needed to induce a stronger relationship between the two. 
Later on it is also observed that deep neural network ensembles can be used for uncertainty estimation~\cite{lakshminarayanan2017simple}, where the average predicted probability estimated from the ensemble outputs can lead to a well-calibrated uncertainty estimation.
Recent advances in the field include diversifying the sub-models with varitional information bottleneck and diversity-inducing adversarial loss to further improve the ensemble accuracy~\cite{sinha2020dibs}, and resolving the scalability issue of deep ensembles or Bayesian neural networks by reducing the memory and computation cost with shared-weight across sub-models and rank-1 parameterization~\cite{wen2020batchensemble,dusenberry2020efficient}.
As the contribution of our work is mainly addressing the adversarial robustness issues, our training algorithm is orthogonal to these previous works. Yet we believe it would be a promising direction to combine our algorithm with other newly-proposed ensemble training methods, leading to a robust ensemble with higher accuracy and less memory and computation cost.

\paragraph{Ensemble of diverse sub-models for robustness.}
%Inspired by the success of building high accuracy ensemble with weak learners demonstrated in traditional bagging and boosting research, previous works also try to train an ensemble of small CNN models to try improving the robustness while preserving clean accuracy. 
Given the success of ensemble methods, a recent line of work investigates improving the robustness of an ensemble of small sub-models (especially against transfer adversarial attacks).
Such robust ensembles can be obtained not only by combining individually robust sub-models but also by eliminating adversarial vulnerabilities shared by different sub-models, be they robust or non-robust, so that attacks cannot transfer between the sub-models within the ensemble.
%Shared adversarial vulnerabilities can arise from adversarial gradient alignment or by exploiting shared adversarial features, though we do not claim these are mutually exclusive. 
Several works attempt to promote diversity in internal representations or outputs across sub-models to serve as a mechanism to limit adversarial transferability and improve ensemble robustness.
Pang et al.~\cite{pang2019improving} propose the ADP regularizer, which forces different sub-models to have high diversity in the non-maximal predictions. Kariyappa et al.~\cite{kariyappa2019improving} reduce the overlap between ``Adversarial Subspaces''~\cite{tramer2017space} of different sub-models by maximizing the cosine distance between each sub-model's gradient w.r.t. the input.
Although the ideas behind these methods are intuitive for improving sub-model diversity, these diversity metrics do not in practice align well with diversifying the adversarial vulnerability shared by different sub-models. Thus training the ensemble with these diversity metrics does not lead to satisfying robustness against transferability between sub-models, and consequently the resulted ensemble is still highly non-robust~\cite{tramer2020adaptive}. 
An ensemble diversity metric that can effectively lead to low attack transferability and high overall robustness is still lacking.

% Comments: not sure if we introduce non-robust feature here, or we just mention it in the beginning of method section

\section{Method}
\label{sec:method}
%In this section, we develop our proposed diversity metric and the \ours{} training algorithm for promoting ensemble diversity.
%In the following, in Section~\ref{ssec:div} we first identify the adversarial vulnerability through distilling non-robust features from a CNN model and define our vulnerability diversity metric. The objective of vulnerability diversification is then formulated based on this diversity metric in Section~\ref{ssec:train}. The detailed algorithm for training an ensemble with \ours{} is provided in Section~\ref{ssec:alg}. 

\subsection{Vulnerability diversity metric}
\label{ssec:div}

%As recently observed by Ilyas et al., non-robust features that are incomprehensible to human yet suffice for neural network classification widely exist in standard image datasets. 
%As these non-robust features are highly sensitive to additive noises, utilizing such features will lead to adversarial vulnerability in a CNN model~\cite{ilyas2019adversarial}.
%If two CNN models captures a similar set of non-robust features during the training process, adversarial attack generated on one of them will be easily transferable to the other. 
A recent study~\cite{ilyas2019adversarial} reveals that non-robust features captured by deep learning models are highly sensitive to additive noise, which is the main cause of adversarial vulnerability in CNN models.
Based on this observation, we propose to isolate the vulnerability of CNN models based on their distilled non-robust features.
%, so that the diversity will directly linked to the attack transferability between the models. 
Let us take a CNN model $f_i$ trained on dataset $D$ as an example. 
We consider a \textit{target} input-label pair $(x,y)\in D$ and another randomly-chosen independent \textit{source} pair $(x_s,y_s)\in D$.
Corresponding to the source image $x_s$, the distilled feature of the input image $x$ by the $l$-th layer of $f_i$ can be approximated with the \textit{feature distillation} objective as~\cite{ilyas2019adversarial}:
\begin{equation}
\label{equ:dis}
x'_{f_i^l}(x,x_s) = \operatorname*{argmin}_z \left\| f_i^l(z)-f_i^l(x) \right\|_2^2,~s.t.\ ||z-x_s||_\infty \leq \epsilon,
\end{equation}
where $f_i^l(\cdot)$ denotes the output before the activation (e.g. ReLU) of the $l$-th hidden layer. 
This constrained optimization objective can then be optimized with PGD~\cite{madry2018towards}.
The distilled feature is expected to be visually similar to $x_s$ rather than $x$ but classified as the target class $y$ since the same feature will be extracted from $x'_{f_i^l}$ and $x$ by $f_i$. 
Such misalignment between the visual similarity and the classification result shows that $x'_{f_i^l}$ reflects the adversarial vulnerability of $f_i$ when classifying $x$.
%Such a misalignment between the visual similarity and the classification result is because the distilled $x'$ is classified entirely based on non-robust features learned from $x$.
%As can be seen that $x'$ demonstrates the adversarial vulnerability of $f_A$ on classifying $x$.
%Similarly, another non-robust feature $x'_{f_B^l}$ can be distilled from model $f_B$.
Therefore, we define the \textit{vulnerability diversity} between two models $f_i$ and $f_j$ as:
\begin{equation}
    \label{equ:div}
    d(f_i,f_j):=\frac{1}{2} \mathbbm{E}_{(x,y),(x_s,y_s),l}\left[\mathcal{L}_{f_i}\left(x'_{f_j^l}(x,x_s),y\right)+\mathcal{L}_{f_j}\left(x'_{f_i^l}(x,x_s),y\right)\right].
\end{equation}
% Note: maybe we should introduce the expectation over layers in next section (in the diverse training objective)
% Actually we can just define this asymmetrically, only classify the distilled feature of f1 with f2
Here $\mathcal{L}_f (x,y)$ denotes the cross-entropy loss of model $f$ for an input-label pair $(x,y)$. 
The expectation is taken over the independent uniformly random choices of $(x,y)\in D$, $(x_s,y_s)\in D$, and layer $l$ of models $f_i$ and $f_j$. 
Since the distilled feature has the same dimension as input images, this formulation can be evaluated on models with arbitrary architectures trained on the same dataset.
As $x'_{f_i^l}(x,x_s)$ is visually uncorrelated with $y$, the cross entropy loss $\mathcal{L}_{f_j}(x'_{f_i^l}(x,x_s),y)$ is small only if $f_j$'s vulnerability on $x$'s non-robust features overlaps with that of $f_i$, and vice versa. 
So the formulation in Equation~(\ref{equ:div}) effectively measures the vulnerability overlap between the two models.
%{\RED So the higher the cross entropy loss is, the less overlapping between the vulnerability of the two models.}
Note that the feature distillation process in Equation~(\ref{equ:dis}) can be considered a special case of generating an adversarial example from source image $x_s$ with target label $y$. 
The diversity defined in Equation~(\ref{equ:div}) therefore corresponds to the attack success rate when transferring adversarial examples between the two models in the same way as training cross-entropy loss corresponds to training accuracy. 
%Note: Mention Nate's CVPR attack here?

\subsection{Vulnerability diversification objective}
\label{ssec:train}
As adversarial attacks are less likely to transfer between models with high vulnerability diversity, we propose to apply the metric defined in Equation~(\ref{equ:div}) as an objective during the ensemble training to induce diverse sub-models and block transfer attacks. 
Equation~(\ref{equ:old}) shows a straightforward way to incorporate the diversity metric into the training objective, where for each sub-model $f_i$, the diversity between itself and all other sub-models $f_j$ in the ensemble is maximized when minimizing the original cross-entropy loss: 
\begin{equation}
    \label{equ:old}
    \min_{f_i} \mathbbm{E}_{(x,y)}[\mathcal{L}_{f_i}(x,y)] - \alpha\sum_{j\neq i}d(f_i,f_j).
\end{equation}
As the formulation of $d(f_i,f_j)$ has no upper bound, directly maximizing it may ultimately lead to divergence.
Thus, we revise the training objective as
\begin{equation}
    \label{equ:alt}
    \min_{f_i} \mathbbm{E}_{(x,y)}\left[\mathcal{L}_{f_i}(x,y) + \alpha \sum_{j\neq i}\mathbbm{E}_{(x_s,y_s),l}\left[\mathcal{L}_{f_i}\left(x'_{f_j^l}(x,x_s),y_s\right)\right]\right],
\end{equation}
which has a stronger bound than Equation~(\ref{equ:old}).
The new objective not only encourages the increase of vulnerability diversity but also facilitates the correct classification of the distilled image as $y_s$. 
As such, the objective is well-posed and can be effectively optimized. 

Furthermore, it should be noted that minimizing $\mathcal{L}_{f_i}(x'_{f_j^l}(x,x_s),y_s)$ can effectively contribute to the minimization of $\mathcal{L}_{f_i}(x_s,y_s)$  as the distilled image $x'_{f_j^l}(x,x_s)$ is close to the clean image $x_s$. 
Previous adversarial training research~\cite{madry2018towards,Xie2020Intriguing} also show that it is not necessary to include the clean data loss in the objective. %is not necessary. 
So we further simplify Equation~(\ref{equ:alt}) to %the objective formulated in Equation~(\ref{equ:ensemble}), which we use in practice for training each individual sub-model in \ours.
\begin{equation}
    \label{equ:ensemble}
    \min_{f_i} \mathbbm{E}_{(x,y),(x_s,y_s),l} \sum_{j\neq i}\mathcal{L}_{f_i}(x'_{f_j^l}(x,x_s),y_s),
\end{equation}
which is adopted for training individual sub-models in \ours.
The objective in Equation~(\ref{equ:ensemble}) can be understood as training sub-model $f_i$ with the adversarial examples generated for other sub-models. 
However, \ours{} is fundamentally different from adversarial training. 
Adversarial training process constantly trains a model on white-box attacks against itself and forces the model to capture the robust feature of the dataset. 
In \ours{}, Equation~(\ref{equ:ensemble}) can be minimized if $f_i$ utilizes a different set of features from other sub-models, including non-robust features. %not necessarily capturing the robust feature.
As non-robust features are distributed more commonly in dataset than robust features~\cite{ilyas2019adversarial}, 
capturing and integrating some non-robust features allows \ours{} to reach higher clean accuracy compared to adversarial training.
%{\RED our objective is easier to be learnt comparing to adversarial training. or it is easier to learn our objective than that of adversarial training. [what is the criteria of learning easily? converge faster?]}
Our training process should also be distinguished from that of Tramer et al.~\cite{tramer2017ensemble}, %use a similar objective which 
which trains a single model with adversarial examples transferred from an ensemble of static pretrained sub-models for improving robustness. 
In \ours{}, all the sub-models in the ensemble are being optimized with Equation~(\ref{equ:ensemble}) in a round-robin fashion. 
The procedure dynamically maximizes the diversity of every pair of sub-models, rather than forcing only a single model away from static pretrained sub-models.
The entire training process of \ours{} is elaborated in Section~\ref{ssec:alg}.

\subsection{\ours{} training routine}
\label{ssec:alg}

% A figure for the overall training process?
%\begin{figure}[tb]
%\centering
%    \captionsetup{width=0.9\linewidth}
%		%\captionsetup{justification=centering}
%		\includegraphics[width=1.0\linewidth]{bit.png}
%	\caption{Example of DNN training under bit-level representation with precision $n=3$. (a) Conversion from floating-point weight $W$ to bit-level representation; (b) Training bit-level model weight with STE.}
%	\label{fig:bit}
%	\vspace{-10pt}
%\end{figure}

\begin{algorithm}
\footnotesize
	\caption{\ours{} training routine for a $N$-sub-model ensemble.} 
	\label{alg}
	\begin{algorithmic}[1]
	    \State \texttt{\# initialization and pretraining}
	    \For {$i=1,\ldots,N$}
		    \State Randomly initialize sub-model $f_i$
		    \State Pretrain $f_i$ with clean dataset
	    \EndFor
	    \State \texttt{\# round-robin feature diversification}
		\For {$e=1,\ldots,E$}
		    \State Uniformly randomly choose layer $l$ for feature distillation 
			\For {$b=1,\ldots,B$}
			    \State $(X,Y)\leftarrow$ get batched input-label pairs
			    \State $(X_s,Y_s)\leftarrow$ uniformly sample batched source input-label pairs
%			    \State
			    \State \texttt{\# get distilled batch for each model}
			    \For {$i=1,\ldots,N$}
			        \State $X'_i := x'_{f_i^l}(X,X_s)\leftarrow$ non-robust feature distillation with Equation~(\ref{equ:dis})
			    \EndFor
%			    \State
			    \State \texttt{\# calculate loss and perform SGD update for all sub-models}
			    \For {$i=1,\ldots,N$}
			        \State $\nabla_{f_i}\leftarrow\nabla[\sum_{j\neq i}\mathcal{L}_{f_i}(f_i(X'_j),Y_s)]$
			        \State
			        $f_i\leftarrow f_i-lr\cdot\nabla_{f_i}$
			    \EndFor
			\EndFor
		\EndFor
	\end{algorithmic} 
\end{algorithm}

Algorithm \ref{alg} shows the pseudo-code for training an ensemble of $N$ sub-models. %$f_1, \dots, f_N$. 
%As the diverse training objective is based on the non-robust feature captured by each sub-model, 
We first randomly initialize and pre-train all the sub-models based on the clean dataset so that their feature spaces will be useful and we do not waste time diversifying irrelevant features. 
Though we show in \textbf{Appendix \ref{ap:pre-training}} that training \ours{} from scratch can also lead to better results than other methods.
Then for each batch of training data during the diverse training phase, we randomly sample another batch of source data and use them to distill the non-robust feature following the objective of Equation~(\ref{equ:dis}). 
A PGD optimization scheme is applied in the feature distillation process.
Round-robin training is then employed wherein a single stochastic gradient descent step is performed on each sub-model with the distilled images from all other sub-models and their source labels, as stated in the objective in Equation~(\ref{equ:ensemble}).
This training process is performed on all $B$ batches of training data and repeated for $E$ epochs. 
The layer $l$ for the feature distillation is randomly chosen in each epoch to avoid overfitting to the features of a particular layer. An ablation study along with discussion on this choice is given in \textbf{Appendix \ref{ap:layer_selection}}.
This training routine can effectively increase the vulnerability diversity between each pair of sub-models within the ensemble and block within-ensemble transfer attacks.
Consequently, the overall black-box robustness of the ensemble improves.

DVERGE induces a similar training complexity as adversarial training does.
Both of these methods need extra back propagations to either distill non-robust features or find adversarial examples.
However, DVERGE uses only intermediate features rather than final outputs for distillation so it is marginally faster than adversarial training.
Detailed comparison of the training complexity of \ours{} vs. previous methods can be found in \textbf{Appendix~\ref{ap:train_details}}.
% Note: do we mention adding adversarial training here? Or only mention it in the experiment section?

\section{Experimental results}
\label{sec:exp}
%In this section, we evaluate the proposed training algorithm in a variety of scenarios and compare its performance against recent, related innovations.

\subsection{Setup}
%We evaluate \ours{} on the CIFAR-10 dataset~\cite{krizhevsky2009learning}.
We compare \ours{} with various counterparts, including \textit{Baseline} which trains an ensemble in a standard way and
two previous robust ensemble training methods: \textit{ADP}~\cite{pang2019improving} and \textit{GAL}~\cite{kariyappa2019improving}.
%achieve ensemble robustness via diversity by incorporating a regularizer into the training objective.
%Specifically, ADP encourages sub-models to make diverse non-maximal predictions, and GAL forces direction misalignment among the gradients of loss w.r.t. the inputs on sub-models.
For a fair comparison, %with~\cite{pang2019improving,kariyappa2019improving}, 
we use ResNet-20~\cite{he2016deep} as sub-models and average the output probabilities after the soft-max layer of each sub-model to yield the final predictions of ensembles.
All the evaluations are performed on the CIFAR-10 dataset~\cite{krizhevsky2009learning}.
Training configuration details can be found in \textbf{Appendix \ref{ap:train_details}}.
For \ours{}, we use PGD with momentum~\cite{dong2018boosting} to perform the feature distillation in Equation~(\ref{equ:dis}). 
We conduct 10 steps of gradient descent during feature distillation with a step size of $\epsilon/10$. 
The $\epsilon$ used for each ensemble size to achieve the results in this section was empirically chosen for the highest diversity and lowest transferability, such that
$\epsilon=$ 0.07, 0.05, 0.05 for ensembles with 3, 5, and 8 sub-models, respectively. 
Analysis on the effect of $\epsilon$ is given in \textbf{Appendix \ref{ap:epsilon}}.
Codes for reproducing the experiments are available at \url{https://github.com/zjysteven/DVERGE}.
%{\color{red} where the key take-away is that larger $\epsilon$ can induce higher diversity and stronger black-box transfer robustness, but can sometimes ...} 
%In the following discussion, we will report the results under $\epsilon$ of 0.07, 0.05, and 0.05 for ensembles with 3, 5, and 8 sub-models respectively. Detailed analysis on the effect of $\epsilon$ is given in \textbf{Appendix \ref{ap:epsilon}}.  

%\sout{In Section~\ref{sec:4.2}, we show that \ours{} can successfully induce diverse vulnerability and suppress the attack transferability within an ensemble. Then, we demonstrate the superior black-box and white-box robustness over other ensemble methods in Section~\ref{sec:4.3}. Further robustness improvement can be obtained by combining \ours{} with adversarial training~\cite{madry2018towards}, which is discussed in Section~\ref{sec:4.4}.}

%We also include the results of adversarial training (referred as \textit{AdvT}) using common configuration \cite{madry2018towards}. 
%Again, more details can be found in \textbf{Appendix \ref{ap:train_details}}. {\color{red} Note, in this work we aim to enhance the robustness of the ensemble as a whole against black-box transfer-based adversarial attacks by diversifying the vulnerability of sub-models rather than eliminating the vulnerability like what adversarial training does. With that being said, the goal of \ours{} is actually orthogonal to that of adversarial training. Therefore, we do not really intend to compare \ours{} with adversarial training.}

\subsection{Diversity and transferability within the ensemble}
\label{sec:4.2}

The objective of \ours{} is to guide sub-models to capture diverse non-robust features and minimize the vulnerability overlap between sub-models, thereby reducing the attack transferability within the ensemble.
To validate our method, we measure diversity and transferability by randomly picking 1,000 test samples on which all sub-models initially give correct predictions. %on these samples.
%For each sub-model $f_i$, we perform 10-step feature distillations on selected samples with a step size of $\epsilon/10$, and feed distilled images to other sub-models $f_j$. 
We compute the pair-wise diversity as the expected
%(w.r.t. the layer and source-target image pair) 
cross-entropy loss formulated in Equation~(\ref{equ:div}), which is further averaged across all pairs of sub-models to obtain the diversity measurement of the whole ensemble.
When measuring the transferability, we generate untargeted adversarial examples using 50-step PGD with a step size of $\epsilon/5$ and five random starts.
The transferability is measured by the attack success rate which counts any misclassification as a success.
Similar to diversity, the averaged pair-wise attack success rate is used to indicate the level of transferability within the ensemble.

\begin{figure}[tb]
\centering
%   \rule{12cm}{3cm}
\captionsetup{width=0.9\linewidth}
  \includegraphics[width=.75\linewidth]{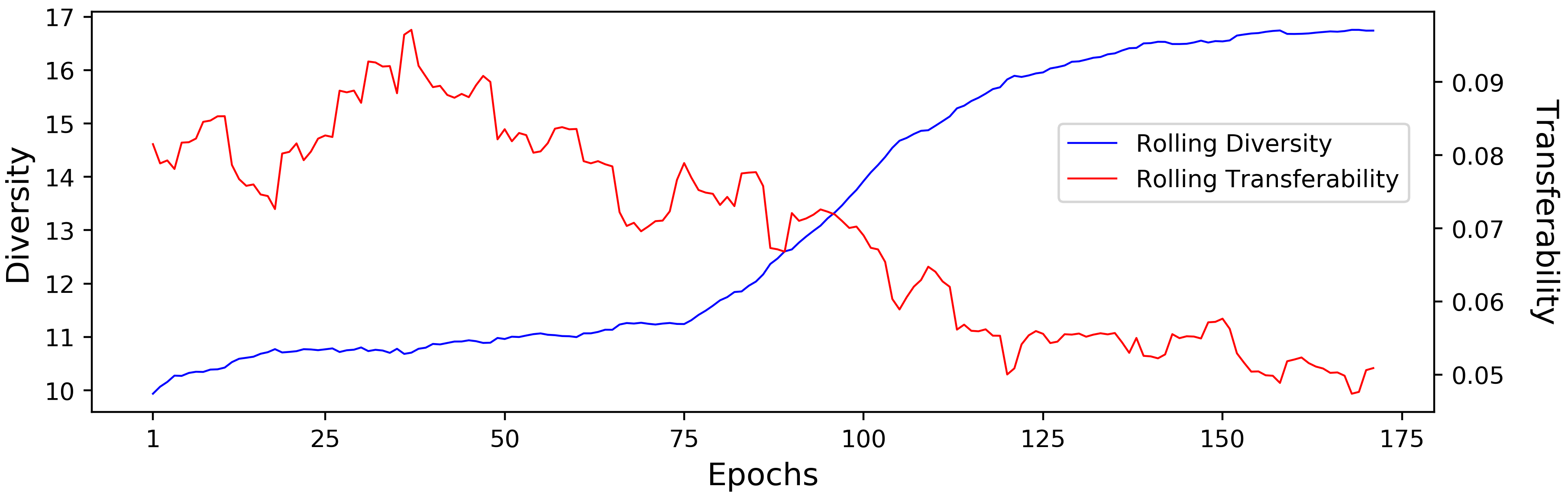}
  \caption{The trend of diversity and transferability during the training of \ours{}. The results are rolling averaged with a window size of 30.}
  % Need to add how the axis are chosen
  \label{fig:div_tran_curve}
  \vspace{-10pt}
\end{figure}

First, let's take a look at how the diversity and transferability within the ensemble changes during the training process of \ours{}.
Figure~\ref{fig:div_tran_curve} shows the result of an ensemble of three sub-models. 
The diversity is evaluated using the same $\epsilon$ of 0.07 as during the training, and the transferability is measured using the standard $\epsilon$ of 0.03 ($\approx8/255$) for adversarial attacks on CIFAR-10~\cite{madry2018towards}. 
The figure clearly shows that the diversity increases while the transferability decreases as the training proceeds. 
This trend empirically validates that minimizing the \ours{} objective can effectively lead to a higher diversity and a lower adversarial transferability within an ensemble.
%Note that as defined in Equation~(\ref{equ:ensemble}), the minimization of the proposed training objective will directly lead to a higher diversity. Therefore, the trend of two quantities changing in the opposite direction indicates the effectiveness of \ours{} to lower the adversarial transferability within an ensemble. 

%Specifically, we randomly pick 1,000 samples from the test set, on which all sub-models can initially give correct predictions. 
%For each sub-model $f_i$, we perform 10-step feature distillation on selected samples with step size of $\epsilon/10$, and feed distilled images to other sub-models $f_j$. 
%The expected cross-entropy loss in Equation~(\ref{equ:ensemble}) is the diversity. %$\mathbbm{E}_{(x,y),(x_s,y_s),l}\left[\mathcal{L}_{f_j}\left(x'_{f_i^l}(x,x_s),y\right)\right]$. 
%When measuring transferability, we generate untargeted adversarial examples using 50-step PGD with an step size of $\epsilon/5$. The attack is equipped with 5 random starts. 
%The transferability is measured by the attack success rate, which counts any misclassification into success. 
%Our experiment shows that 50 steps is sufficient for PGD to converge (\textbf{Appendix \ref{ap:convergence}}). %Note, increasing PGD iterations to 5,000 would only raise the attack success rate by no more than 1\%, so here 50 steps is sufficient for PGD to converge.

\begin{figure}[tb]
\centering
\captionsetup{width=0.9\linewidth}
  \includegraphics[width=.85\linewidth]{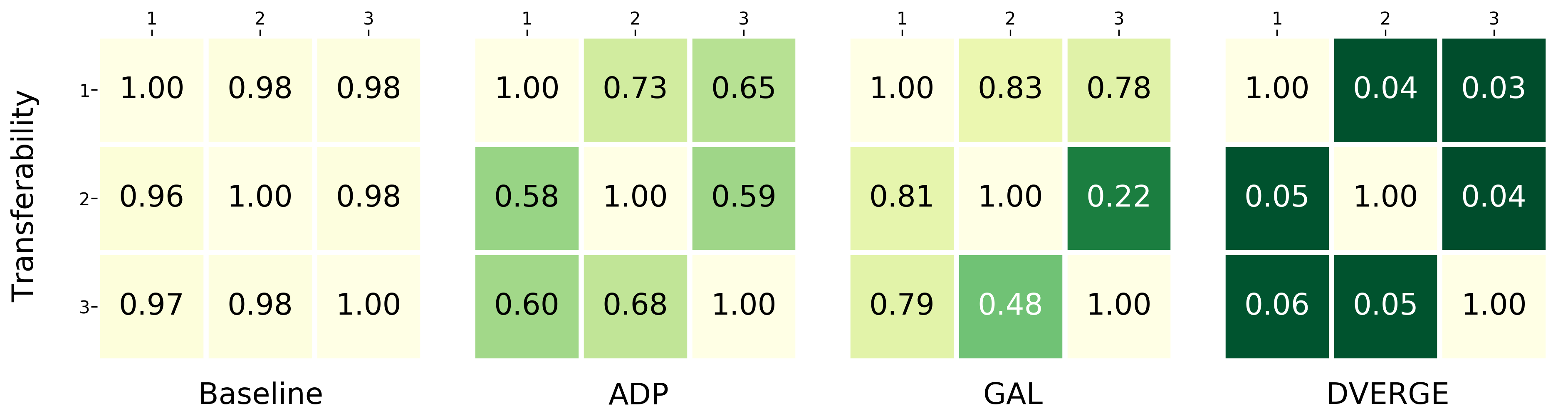}
  \caption{Pair-wise transferability (in the form of attack success rate) among sub-models for different ensemble methods.} %Diagonal numbers represent the white-box attack success rate of each sub-model.}
  \label{fig:transf}
  \vspace{-10pt}
\end{figure}

Figure~\ref{fig:transf} presents the pair-wise transferability of an ensemble with three sub-models tested under the same $\epsilon$ as aforementioned. 
Results for the ensembles composed of more sub-models and for other testing $\epsilon$ are reported in \textbf{Appendix \ref{ap:epsilon}} and \textbf{Appendix \ref{ap:trans_test_eps}}, respectively. 
%In every heatmap, 
The number at the intersection of the $i$-th row and $j$-th column represents the transfer success rate of the adversarial examples generated from the $i$-th sub-model and tested on the $j$-th sub-model.
When $i=j$, the number becomes the white-box attack success rate.
Larger off-diagonal numbers indicate greater transferability across sub-models.
Compared with other ensemble methods, \ours{} suppresses the transferability to a much lower level; 
among all adversarial examples that successfully break one sub-model, only 3-6\% of them could lead to misclassification on other sub-models. 
Although ADP and GAL also strive to improve the diversity for better robustness, they cannot effectively block adversarial transfer. 
ADP exhibits transference of 60\% to 70\% of attacks between sub-models.
When it comes to GAL, two out of the three sub-models are still extremely vulnerable to each other, where more than 80\% of adversarial examples can successfully transfer between the first and second sub-models. 
Our evaluation demonstrates that stopping attack transfers is not trivial and applying an appropriate diversification metric is crucial. 
Therefore, we advocate the use of \ours{} as a more effective means for mitigating the attack transferability within an ensemble.

\subsection{Robustness of the ensemble}
\label{sec:4.3}

\begin{figure}[tb]%[!thbp]
\centering
\captionsetup{width=0.9\linewidth}
  \includegraphics[width=\linewidth]{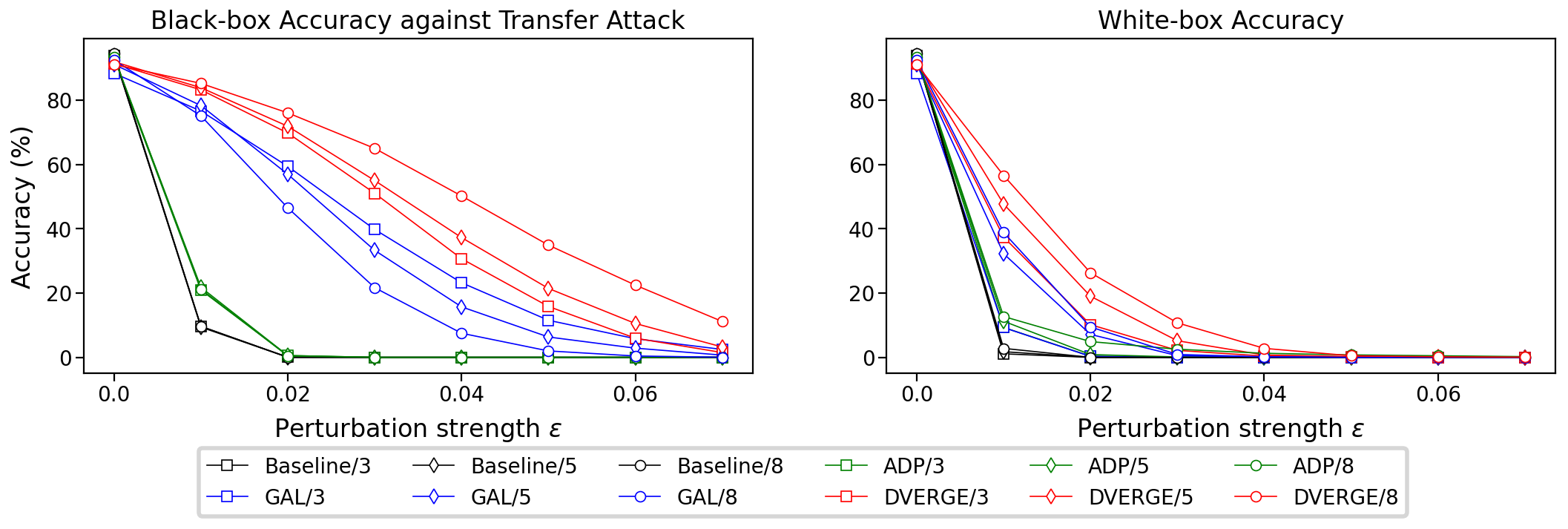}
  \caption{Robustness results for different ensemble methods. The number after the slash stands for the number of sub-models.}
  \label{fig:rob}
  \vspace{-10pt}
\end{figure}

We evaluate the robustness of ensembles under two threat models: black-box transfer adversary, where the attackers cannot access the model parameters and rely on surrogate models to generate transferable adversarial examples,
%{\color{red} We do not aim to defend against query-based black-box attacks in this work.}
and white-box adversary, where the attackers have the full access of everything of the model. 
Under the black-box scenario, we use hold-out baseline ensembles with 3, 5, and 8 ResNet-20 sub-models as the surrogate models. 
%We generate adversarial examples on each of these ensembles and use the collection of all adversarial images to attack target ensembles. 
A more challenging setting considers an attacker fully aware of the defense such that the surrogate ensemble is trained with the exact technique.
The results under this setting can be seen in \textbf{Appendix~\ref{ap:rob_number}}.
We use three attack methodologies:  
(1) PGD with momentum~\cite{dong2018boosting} with three random starts.
%, which has been shown to boost transferability \cite{dong2018boosting}. 
%Three random starts are incorporated.
(2) M-DI\textsuperscript{2}-
FGSM \cite{xie2019improving}, which randomly resizes and pads the image in each step of attack generation.
(3) SGM \cite{wu2020skip}, which adds weight to the gradient through the skip connections of ResNets. 
The latter two attacks are essentially two stronger black-box transfer attacks that can better expose the attack transferability between models.
For more details, we refer the reader to the attacks' respective papers. 
We run each attack for 100 iterations with the step size of $\epsilon/5$.
Other than using the cross-entropy loss, we also generate adversarial examples with CW loss \cite{carlini2017towards} since it can also help with the transfer.
As a result, in total, each sample will have 3 (surrogate models)\x 5 (PGD with 3 random starts plus 2 other attacks)\x 2 (loss functions) = 30 adversarial counterparts. 
The black-box accuracy is reported in a \textit{all-or-nothing} fashion: We say the model is accurate on one sample only if all of its 30 adversarial versions are correctly classified by the model.
We adopt such a powerful adversary and a strict criteria to give a tighter upper bound of the robustness against black-box transfer attacks.
Under the white-box scenario, we use 50-step PGD with five random starts and the step size of $\epsilon/5$ to attack ensembles.
Our results in \textbf{Appendix~\ref{ap:convergence}} prove that we have applied sufficient steps for attacks to converge.

%Again, we confirm the convergence of the crafted attacks in \textbf{Appendix \ref{ap:convergence}}. %Note again that using more iterations, say, 5,000, only decreases the accuracy by 0.2\%, so we confirm that 1,000 iterations is sufficient for PGD to converge in this case.

%against black-box transfer attacks, the scenario under which we claim that \ours{} could introduce robustness into the ensemble. More specifically, the attacker is assumed to generate adversarial examples on surrogate models and then transfer them to the target ensemble. We consider three attacks here.  Three hold-out standard ensembles with 3, 5, and 8 ResNet-20s serve as the candidates of surrogate models, which means we generate adversarial examples on each of these ensembles and use the collection of adversarial images to attack. In total, each sample will have 3(surrogate models)\x 2(loss functions)\x 5(PGD with 3 random starts plus 2 other attacks)=30 adversarial counterparts. The black-box accuracies are reported in a \textit{all-or-nothing} fashion: We say the model is accurate on one sample only if all of its 30 adversarial versions are correctly classified by the model. Again, we confirm the convergence of the crafted attacks in \textbf{Appendix \ref{ap:convergence}}. %Note again that using more iterations, say, 5,000, only decreases the accuracy by 0.2\%, so we confirm that 1,000 iterations is sufficient for PGD to converge in this case.

Evaluated on 1,000 randomly selected test samples, Figure \ref{fig:rob} shows the black-box and white-box robustness of ensembles with various number of sub-models across a wide range of attack budget $\epsilon$. Here we show the averaged results over three independent runs.
We refer the reader to %Table \ref{tab:ap_bbox_result} and \ref{tab:ap_wbox_result} in
\textbf{Appendix~\ref{ap:rob_number}} for the plots with error bars and numerical results.
\ours{}, even with the least sub-models, outperforms each case of the other methods with higher accuracy in both black-box and white-box settings and achieves comparable clean accuracy.
In addition, robustness improvement can be easily obtained by adding more sub-models into the ensemble when using our method, while such a satisfying trend is less obvious in other methods.
GAL, as the second best performing approach among the four methods, actually shares the same high-level concept as the proposed \ours{} algorithm.
They both aim at diversifying the vulnerabilities shared by the sub-models.
%The difference lies in that GAL measures diversity with the misalignment of gradient directions and uses a diversity regularizer to reduce the vulnerability overlap, while \ours{} defines diversity from the perspective of non-robust features and adapts the training routine to boost the diversity.
The difference lies in the fact that GAL considers using the adversarial gradient directions to evaluate the vulnerability of CNN models whereas \ours{} identifies the vulnerability in a model by distilling the learnt non-robust features.
Results from both Figure \ref{fig:transf} and Figure \ref{fig:rob} suggest that our approach is a more effective realization of the intuition of identifying and diversifying adversarial vulnerability.

\subsection{\ours{} with adversarial training}
\label{sec:4.4}

\begin{figure}[tb]%[!thbp]
\centering
  \includegraphics[width=\linewidth]{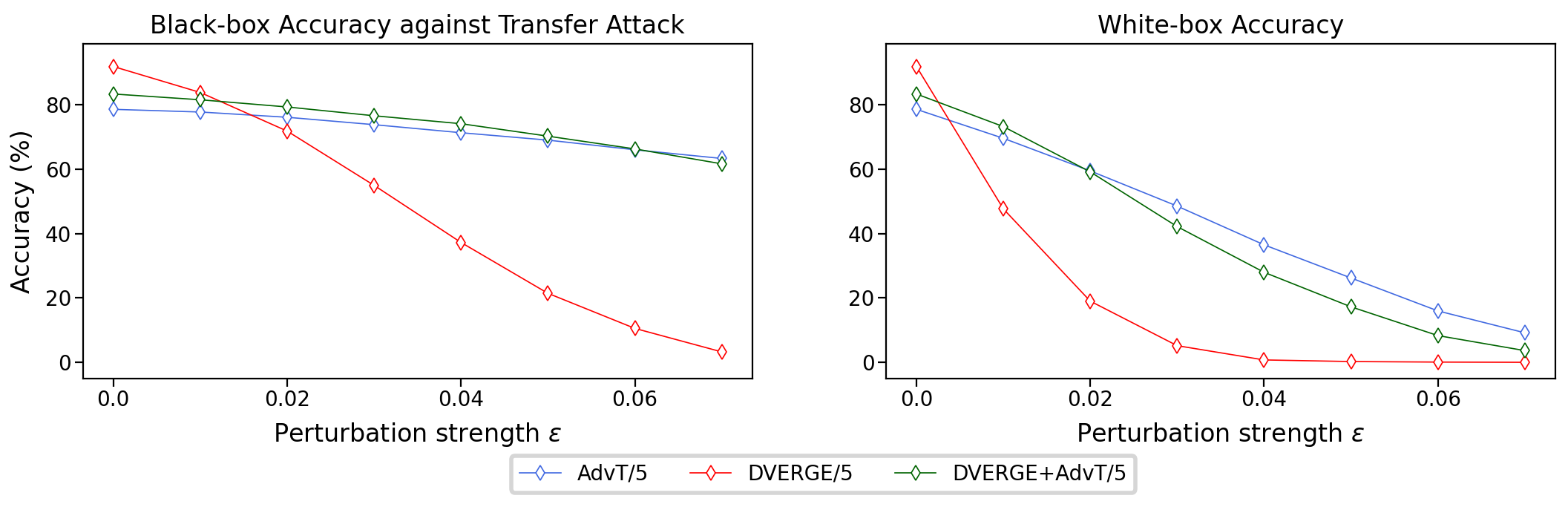}
  \caption{Results for \ours{} combined with adversarial training.}
  \label{fig:rob_ours_adv}
  \vspace{-10pt}
\end{figure}

Although \ours{} achieves the highest robustness among ensemble methods, its robustness against white-box attacks and transfer attacks with a large perturbation strength is still quite low. 
This result is expected because the objective of \ours{} is to diversify the adversarial vulnerability rather than completely eliminate it.
%The analysis in the previous subsection shows that as the attack perturbation strength increases,  the accuracy of all the methods, including \ours{}, drops to a low level, especially under the white-box scenario.
%This is because ensemble methods aim to diversify the adversarial vulnerability rather than eliminating it. 
In other words, vulnerability inevitably exists within sub-models and can be captured by attacks with larger $\epsilon$.
One straightforward way to improve the robustness of ensembles is to augment \ours{} with adversarial training \cite{madry2018towards}. %which is the state-of-the-art defense technique for individual models.
We describe implementation details regarding adversarial training in \textbf{Appendix \ref{ap:train_details}} and the amended objective in \textbf{Appendix \ref{ap:ours_plus_adv}}.

Figure \ref{fig:rob_ours_adv} presents the black-box and white-box accuracy for adversarial training (\textit{AdvT}), \ours{} only (\textit{\ours}) and the combination of the two (\textit{\ours{}+AdvT}) using the same evaluation setting as in Section \ref{sec:4.3}.
Ensembles with 5 sub-models are used here, and the results are averaged over three independent runs. 
More results under different ensemble sizes and plots with error bars can be found in \textbf{Appendix \ref{ap:ours_plus_adv}}.
%See \textbf{Appendix \ref{ap:result}} for more results with different ensemble sizes.
%The results shown in Figure \ref{fig:rob_ours_adv} can be seen as another evidence for the recent finding \cite{ilyas2019adversarial,tsipras2018robustness} that robust features provide robustness yet are harmful for the performance on benign samples, while non-robust features help with the clean accuracy but are extremely vulnerable to adversarial attacks.
The \ours{}+AdvT objective favors the capture of more robust features by the ensemble. %, which leads to a higher robustness against both white-box and transfer attacks comparing to training with \ours{} only.
Compared to AdvT, \ours{}+AdvT encourages the ensemble to learn diverse non-robust features alongside robust features, leading to a higher clean accuracy and higher robustness against transfer attacks.
In the meantime, no matter which objective is applied, the overall learning capacity of the ensemble remains the same. %remains the same under all three objectives, so 
That is, learning more robust features will leave less capacity in the ensemble to capture diverse non-robust features, and vice versa. 
Forcing the inclusion of robust features causes \ours{}+AdvT to sacrifice the accuracy on clean examples comparing to performing \ours{} only. 
Learning diverse non-robust features harms \ours{}+AdvT's robustness against white-box attacks with larger perturbations compared to AdvT alone.
%Since \ours{}+AdvT embraces both non-robust features and robust features from each single method, it yields better clean accuracy than adversarial training and stronger robustness than \ours{}.
These results can be seen as an evidence for the recent findings in~\cite{ilyas2019adversarial,tsipras2018robustness} regarding the tradeoff between clean accuracy and robustness.
\ours{}+AdvT can effectively explore such tradeoff by changing the ratio between the two objectives, which is further illustrated in \textbf{Appendix~\ref{ap:ours_plus_adv}}.
%In fact, \ours{}+AdvT naturally characterizes the trade-off between clean accuracy and robustness, which can be observed by simply changing the ensemble size (see \textbf{Appendix \ref{ap:result}}).
%Furthermore, with the help of diverse non-robust features, \ours{}+AdvT is more robust than adversarial training against black-box transfer attacks.
%\ours{} degrades the robustness achieved by adversarial training against white-box attacks with larger perturbation, though, where non-robust features can be easily exploited and compromised. 

\section{Conclusions}
\label{sec:con}
In this work we propose \ours{}, a CNN ensemble training method that isolates and diversifies the adversarial vulnerability in each sub-model to improve the overall robustness against transfer attacks without significantly reducing clean accuracy.
We show that adversarial diversity in a CNN model can be successfully characterized by distilled non-robust features, from which we can measure the vulnerability diversity between two models.
The diversity metric is further developed into the vulnerability diversification objective used for \ours{} ensemble training.
% Experiment results
We empirically show that training with \ours{} objective can effectively increase the vulnerability diversity between sub-models, thereby blocking attack transferability within the ensemble. 
In this way \ours{} reduces the success rate of transfer attacks between sub-models from more than 60\% achieved by previous ensemble training methods to less than 6\%, which enables ensembles trained with \ours{} to achieve significantly higher robustness against both black-box transfer attacks and white-box attacks compared to previous ensemble training methods. 
The robustness can be further improved with additional sub-models in the ensemble.
We further demonstrate that \ours{} can be augmented with an adversarial training objective, %where the ensemble will be encouraged to learn both robust and diverse non-robust features. This
which enables the ensemble to achieve higher clean accuracy and higher transfer attack robustness compared to adversarial training. %, and achieve higher white-box robustness than training only with the \ours{} objective.
In conclusion, the vulnerability diversity induced by \ours{} training objective can effectively contribute to enhancing the robustness of CNN ensembles while maintaining desirable clean accuracy.
\section*{Broader Impact}
\ours{} hypothetically addresses some black-box adversarial vulnerabilities pervasive across machine learning applications while increasing compute requirements to training models. 
As such methods presented herein suggest potential impacts on the reliability, security, and carbon-footprint of deep-neural-network-based systems. 
The reliability and robustness of machine learning systems are not just a concern for practitioners but also policy makers \cite{hamonrobustness}.

A net increase in carbon production would be considered a negative impact by many researchers in climate-related fields. 
This problem is common to many techniques that modify model training to achieve robustness, including \ours. 
While yet to be examined in the case of \ours, the possibility to mitigate or reduce excessive training burdens through informed hyperparameter selection exists.
Sometimes, though modified training may increase the required computation per model parameter update, the modified method may nevertheless require fewer steps or epochs to achieve desirable results. 
Recent work provides actionable recommendations, such as performing cost-benefit analysis, to determine if efficient downstream adoption is desirable \cite{strubell2018linguistically}.

In both industrial and military applications, practical solutions to vulnerabilities, such as relying on human-AI teaming \cite{danks2020}, are effective but do not address the underlying source of vulnerability and may limit the adoption of machine learning elsewhere. 
Addressing vulnerabilities at the training stage, then, is a desirable capability for positive-impact applications. 
By orthogonally improving only black-box robustness, though, we leave machine learning systems vulnerable to other types of attacks. 
Previous work has shown that white-box knowledge can still be leaked in black-box scenarios \cite{oh2019towards,tramer2016stealing}. 
As such, \ours{} is reliant on adversarial training to defend against white-box attacks and on traditional computer security to maintain system integrity. 
The ultimate interpretation of impact due to improved model reliability and security is not clear-cut, however, as it is highly dependent on the application space. 
This uncertainty is symptomatic of the fact that machine learning is often fundamental by nature and that there is no machine learning technique for improving robustness that can be applied only to positive-impact applications, whatever one's subjective interpretation of ``positive'' may be. 
% In order to provide a balanced perspective, authors are required to include a statement of the potential broader impact of their work, including its ethical aspects and future societal consequences. Authors should take care to discuss both positive and negative outcomes.
% Not counted in the 8-page limit

\section*{Acknowledgments and Disclosure of Funding}

This work is supported by the DARPA HR00111990079 (QED for RML) program.

\medskip

\small

\bibliographystyle{unsrt}
\bibliography{neurips_2019}

\newpage
\normalsize
\appendix
\section{Training and implementation details}
\label{ap:train_details}

We train the baseline ensembles for 200 epochs using SGD with momentum 0.9 and weight decay 0.0001. 
The initial learning rate is 0.1, and we decay it by 10\x~at the 100-th and the 150-th epochs.
Any models pre-trained on the clean dataset can serve as the starting point for the training of \ours{}. 
In our implementation, \ours{} starts from the trained baseline ensembles.
We follow the aforementioned learning rate schedule, but using a carefully-tuned one is likely to bring extra performance gain.
We reproduce ADP \cite{pang2019improving} and GAL \cite{kariyappa2019improving} according to either the released code or the paper with recommended hyperparameters and setups.
Specifically, they both use Adam optimizer \cite{kingma2014adam} with an initial learning rate of 0.001.
Also note that GAL requires the ReLU function to be replaced with leaky ReLU to avoid gradient vanishing.
The other configurations stay the same as those of baseline ensembles.

%We find the recreated version gives numbers which closely match the reported ones when using the same evaluation configuration. 

Ensembles with adversarial training follow the baseline's training setup. We use 10-step PGD with $\epsilon=8/255$ and a step size $\alpha=2/255$ \cite{madry2018towards}.
More specifically, adversarial examples w.r.t. the whole ensemble are generated in each step of the training process and are used to update model parameters.
When combining \ours{} with adversarial training, however, adversarial examples are generated on each sub-model instead of the whole ensemble.
We empirically find these choices help each case achieve its best robustness.

We use 0.5 as the input transformation probability for M-DI\textsuperscript{2}-
FGSM \cite{xie2019improving} and 0.2 as the $\gamma$ for SGM \cite{wu2020skip} when generating these two attacks as recommended by their respective papers.

All models are implemented and trained with PyTorch~\cite{paszke2017automatic} on a single NVIDIA TITAN XP GPU.
Evaluation is performed based on AdverTorch~\cite{ding2019advertorch}. 
As shown in Table~\ref{tab:ap_train_time}, the training of DVERGE is marginally faster than that of adversarial training (AdvT). As they both need extra back propagations to either distill non-robust features or find adversarial examples, DVERGE uses only intermediate features for distillation while adversarial training requires the information back propagated from the final output.
As for previous methods, though ADP requires the least time budget, it does not improve the robustness much as shown in Figure~\ref{fig:rob}. And the significantly improved robustness would be worth the extra training cost of DVERGE over GAL. 
In addition, according to Figure~\ref{fig:div_tran_curve}, DVERGE could reduce the transferability within the ensemble at the very early stage of the training, so later training epochs have the potential to be simplified to a fine-tuning process without diversity loss, which will require much less training time. Further mitigating the computational overhead of DVERGE would be one of our future goals.

\begin{table}[!hbt]
    %\small
    \caption{Training time comparison on a single TITAN XP GPU. All times are evaluated for training a ResNet-20 ensemble with 3 sub-models for 200 epochs.}
    \label{tab:ap_train_time}
    \centering
    \begin{tabular}{c|ccccc}
    \toprule
    Method & Baseline & ADP~\cite{pang2019improving} & GAL~\cite{kariyappa2019improving} & AdvT~\cite{madry2018towards} & \textbf{\ours{}}\\
    \midrule
    Training time (h) & 1.0 & 2.0 & 7.5 & 11.5 & \textbf{10.5} \\
    \bottomrule
    \end{tabular}
\end{table}
\section{Analysis on the training $\epsilon$ of \ours{}}
\label{ap:epsilon}

\begin{table}[!hbt]
    %\small
    \footnotesize
    \caption{The effect of $\epsilon$ on optimizing the feature distillation objective in Equation (\ref{equ:dis}) and the resulting diversity loss measured with Equation (\ref{equ:ensemble}). $f_i$ and $f_j$ are two ResNet-20 models trained in a standard way on CIFAR-10. $x'_{f_i^l}$ is short for $x'_{f_i^l}(x,x_s)$. The step size used for feature distillation is chosen as $\epsilon/\text{\#steps}$. }
    \label{tab:ap_train_eps}
    \centering
    \begin{tabular}{cccc}
    \toprule
    \multirow{2}{*}{$\epsilon$} & \multirow{2}{*}{\#steps} & $\vphantom{\mathbb{E}_{(x,y),(x_s,y_s)}\left\|f_i^l(x'_{f_i^l})-f_i^l(x)\right\|_2} \text{feature distillation objective}$ & diversity loss \\
     & & $\displaystyle \mathop{\mathbb{E}}_{(x,y),(x_s,y_s),l}\left\|f_i^l(x'_{f_i^l})-f_i^l(x)\right\|_2$  &$\displaystyle \mathop{\mathbb{E}}_{(x,y),(x_s,y_s),l}\left[\mathcal{L}_{f_j}(x'_{f_i^l},y_s)\right]$\\
    \midrule
    0.03 & 10 &1.056 &1.726 \\
    %\midrule
    0.05 & 10 &0.793 &3.302 \\
    %\midrule
    0.07 & 10 &0.703 &4.738 \\
    \bottomrule     
    \end{tabular}
\end{table}

\begin{figure}[!htb]
\centering
\captionsetup{width=0.9\linewidth}
  \includegraphics[width=\linewidth]{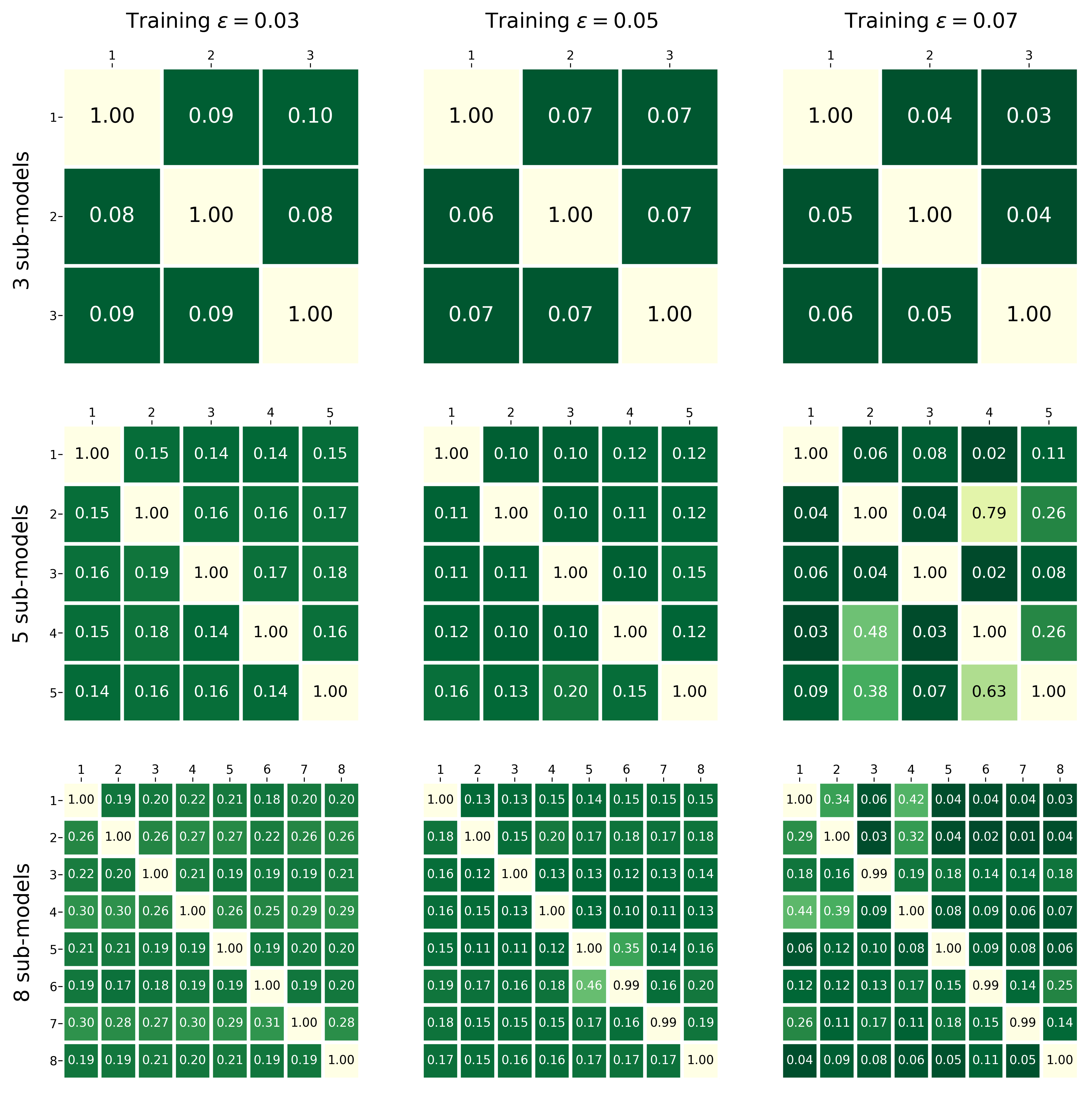}
  \caption{Transferability within \ours{} ensembles trained with different $\epsilon$. The evaluation setting follows that of Figure \ref{fig:transf}, where the attack perturbation strength is 0.03.}
  \label{fig:ap_trans_training_eps_more_submodel}
\end{figure}

One important hyperparameter of \ours{} is the $\epsilon$ used for feature distillation. 
%We leave thorough and rigorous analysis on $\epsilon$ as a future work and
This section provides some initial exploration on the effect of using different $\epsilon$.
We start by looking at how $\epsilon$ affects the optimization of the feature distillation in Equation (\ref{equ:dis}) and the resulted diversity loss in Equation (\ref{equ:ensemble}).
The results evaluated with 1,000 CIFAR-10 testing images on two pre-trained ResNet-20 models are shown in Table \ref{tab:ap_train_eps}. 
We find that a larger $\epsilon$ enables more accurate feature distillation as a smaller distance between the internal representation of the distilled image $x'_{f^l_i}$ and the target image $x$ is achieved.
As a result, the distilled image from model $f_i$ can lead to a higher diversity loss on another model $f_j$, which intuitively encourages the training routine of \ours{} to enforce a greater diversity and therefore a lower transferability between the two models.
We empirically confirm this intuition in Figure \ref{fig:ap_trans_training_eps_more_submodel}, where we vary the training $\epsilon$ and measure the transferability between sub-models.
For instance, when ensembles have three sub-models, increasing $\epsilon$ from 0.03 to 0.07 decreases the transferability from 8\%-10\% to 3\%-6\%.
Interestingly, however, for ensembles with five or eight sub-models, although we do observe a drop in the transferability between most of the sub-model pairs by using a larger $\epsilon$, some pairs of the sub-models remain highly vulnerable to one another.
In particular, observe that when training an ensemble of 5 sub-models with an $\epsilon$ of 0.07, 79\% of adversarial examples from the second sub-model can fool the fourth one and 48\%  of examples transfer in the reverse direction.
We leave a thorough and rigorous analysis of this phenomenon to future work.

Finally, we look at the clean accuracy and robustness achieved by \ours{} ensembles trained with different $\epsilon$. 
In Table \ref{tab:ap_rob_results_under_train_eps}, we observe that training with a larger $\epsilon$ leads to a higher black-box transfer robustness but a lower clean accuracy.
The trend between white-box robustness and $\epsilon$ is not monotonic though, which we suspect is related to the observation that the attack transferability worsens between some pairs of sub-models under a larger training $\epsilon$, as shown in Figure \ref{fig:ap_trans_training_eps_more_submodel}.
As the robustness against white-box attacks is not the main focus of \ours, we will explore the relationship, both qualitatively and quantitatively, between the transferability among sub-models and the achieved white-box robustness of the whole ensemble in the future.

\begin{table}[!hbt]
    %\small
    \footnotesize
    \caption{Robustness of \ours{} ensembles trained with different $\epsilon$. In each table block, we report (clean accuracy)~/~(black-box transfer accuracy under perturbation strength 0.03)~/~(white-box accuracy under perturbation strength 0.01).}
    \label{tab:ap_rob_results_under_train_eps}
    \centering
    \begin{tabular}{|c||c|c|c|}
    \hline
    \diagbox[width=10em]{\#sub-models}{training $\epsilon$}
     &0.03 &0.05 &0.07 \\
    \hline\hline
    3 &92.9\%~/~~~4.2\%~/~22.7\% &92.7\%~/~26.6\%~/~32.3\% 
    &91.4\%~/~53.2\%~/~40.0\% \\\hline
    5 &92.3\%~/~30.5\%~/~43.1\%
    &91.5\%~/~57.2\%~/~48.9\%
    &90.2\%~/~66.5\%~/~42.3\% \\\hline 
    8 &91.3\%~/~42.8\%~/~51.9\%
    &91.1\%~/~63.6\%~/~57.9\%
    &89.2\%~/~71.3\%~/~52.4\% \\\hline 
    \end{tabular}
\end{table}

\section{Additional results}
\label{ap:result}

\subsection{Decision region visualization}
\label{ap:decision_region}

\begin{figure}[!h]
\centering
\captionsetup{width=0.9\linewidth}
  \includegraphics[width=.8\linewidth]{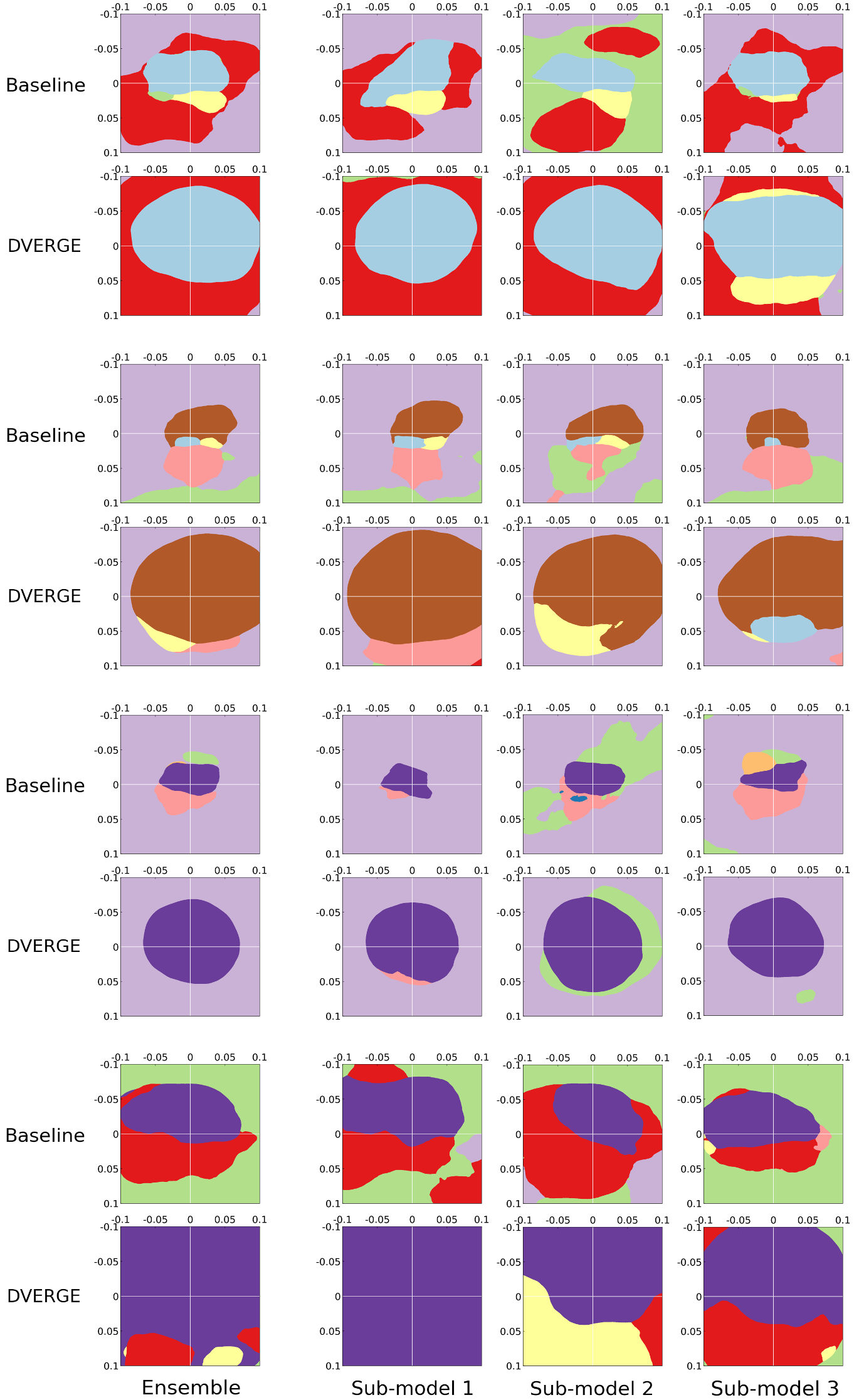}
  \caption{Additional decision region plots of ensembles with 3 ResNet-20 sub-models trained on CIFAR-10. Each pair of two rows is generated with one testing image. The first row is for the baseline ensemble, and the second row is for the \ours{} ensemble. The axes are chosen in the same way as in Figure~\ref{fig:dec}.}
  % Need to add how the axis are chosen
  \label{fig:ap_dec}
\end{figure}

We visualize the decision regions learned by \ours{} ensembles around more testing images from the CIFAR-10 dataset in Figure \ref{fig:ap_dec}.

\subsection{Transferability within the ensemble under different testing $\epsilon$}
\label{ap:trans_test_eps}

\begin{figure}[!htb]
\centering
\captionsetup{width=0.9\linewidth}
  \includegraphics[width=\linewidth]{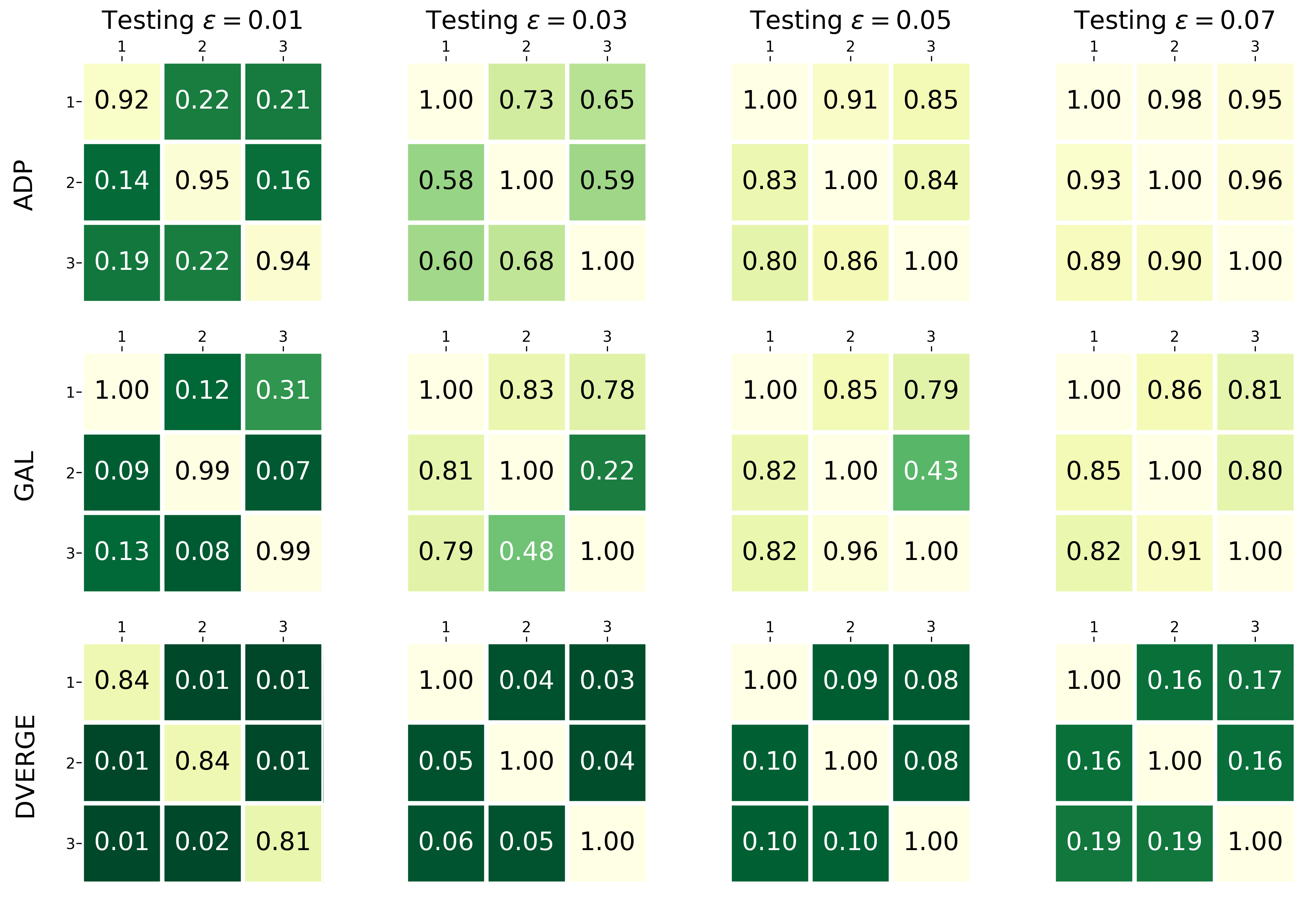}
  \caption{Transferability results under different testing $\epsilon$.}
  % Need to add how the axis are chosen
  \label{fig:ap_trans_test_eps}
\end{figure}

In addition to Figure \ref{fig:transf}, we provide more results of the transferability between sub-models under different attack $\epsilon$ in Figure \ref{fig:ap_trans_test_eps}. 
In all cases, \ours{} achieves the lowest level of transferability among all ensemble methods.

\begin{table}[!hbt]
    %\small
    \footnotesize
    \caption{Accuracy v.s. $\epsilon$ against black-box transfer attacks generated from hold-out baseline ensembles. The number in the first column after the slash is the number of sub-models within the ensemble. The results are averaged over three independent runs.}
    \label{tab:ap_bbox_result}
    \centering
    \begin{tabular}{c|cccccccc}
    \toprule
    $\epsilon$ & clean & 0.01 & 0.02 & 0.03 & 0.04 & 0.05 & 0.06 & 0.07 \\
    \midrule
    \midrule
    baseline/3 & 93.9\%	& 9.6\% &0.1\% &0\% &0\% &0\% &0\% &0\% \\  
    baseline/5 & 93.9\%	& 9.4\% &0\% &0\% &0\% &0\% &0\% &0\% \\ 
    baseline/8 & 94.4\%	& 9.7\% &0\% &0\% &0\% &0\% &0\% &0\% \\ 
    \midrule
    ADP/3 \cite{pang2019improving} & 92.8\% & 20.9\% &0.6\% &0\% &0\% &0\% &0\% &0\% \\ 
    ADP/5 \cite{pang2019improving} & 93.2\% & 21.8\% &0.6\% &0\% &0\% &0\% &0\% &0\% \\ 
    ADP/8 \cite{pang2019improving} & 93.4\% & 21.2\% &0.4\% &0\% &0\% &0\% &0\% &0\% \\ 
    \midrule
    GAL/3 \cite{kariyappa2019improving}
    & 88.3\% & 76.6\% &59.4\% &39.8\% &23.2\% &11.5\% &5.8\% &2.5\% \\ 
    GAL/5 \cite{kariyappa2019improving} & 91.1\% & 78.3\% &56.9\% &33.3\% &15.7\% &6.3\% &2.9\% &0.7\% \\ 
    GAL/8 \cite{kariyappa2019improving} & 92.4\% & 75.1\% &46.5\% &21.7\% &7.5\% &2.0\% &0.4\% &0.0\% \\  
    \midrule
    \ours{}/3 & 91.3\% & 83.2\% &69.7\% &51.0\% &30.7\% &15.8\% &6.0\% &1.5\% \\
    \ours{}/5 & 91.9\% & 83.8\% &71.8\% &55.0\% &37.2\% &21.4\% &10.5\% &3.2\% \\
    \ours{}/8 & 91.1\% & 85.2\% &76.0\% &65.0\% &50.2\% &34.9\% &22.5\% &11.2\% \\
    \midrule
    \midrule
    AdvT/3 \cite{madry2018towards} &77.2\% &76.3\% &74.8\% &73.2\% &70.9\% &68.7\%	&66.0\%	&62.7\% \\
    AdvT/5 \cite{madry2018towards} &78.6\% &77.8\% &76.1\% &73.8\% &71.3\% &69.0\%	&66.0\%	&63.3\% \\
    AdvT/8 \cite{madry2018towards} &79.4\% &78.2\% &76.9\% &74.9\% &72.3\% &69.8\%	&66.8\%	&63.9\% \\
    \midrule
    \ours{}+AdvT/3 & 81.4\% & 79.7\% &77.6\% &75.3\% &72.5\% &69.1\% &66.4\% &62.6\% \\
    \ours{}+AdvT/5 & 83.4\% & 81.6\% &79.3\% &76.6\% &74.1\% &70.3\% &66.3\% &61.7\% \\
    \ours{}+AdvT/8 & 85.0\% & 82.8\% &80.2\% &76.8\% &73.0\% &68.7\% &64.0\% &57.9\% \\
    \bottomrule     
    \end{tabular}
\end{table}

\begin{table}[!htb]
    %\small
    \footnotesize
    \caption{Accuracy v.s. $\epsilon$ against white-box attacks. The number in the first column after the slash is the number of sub-models within the ensemble. The results are averaged over three independent runs.}
    \label{tab:ap_wbox_result}
    \centering
    \begin{tabular}{c|cccccccc}
    \toprule
    $\epsilon$ & clean & 0.01 & 0.02 & 0.03 & 0.04 & 0.05 & 0.06 & 0.07 \\
    \midrule
    \midrule
    baseline/3 & 93.9\%	& 1.1\% &0\% &0\% &0\% &0\% &0\% &0\% \\  
    baseline/5 & 93.9\%	& 1.8\% &0\% &0\% &0\% &0\% &0\% &0\% \\ 
    baseline/8 & 94.4\%	& 2.8\% &0\% &0\% &0\% &0\% &0\% &0\% \\ 
    \midrule
    ADP/3 \cite{pang2019improving} & 92.8\% & 9.3\% &0.4\% &0\% &0\% &0\% &0\% &0\% \\ 
    ADP/5 \cite{pang2019improving} & 93.2\% & 11.2\% &0.9\% &0.1\% &0\% &0\% &0\% &0\% \\ 
    ADP/8 \cite{pang2019improving} & 93.4\% & 12.7\% &4.9\% &2.5\% &1.3\% &0.8\% &0.5\% &0.2\% \\ 
    \midrule
    GAL/3 \cite{kariyappa2019improving}
    & 88.3\% & 9.3\% &0.3\% &0\% &0\% &0\% &0\% &0\% \\ 
    GAL/5 \cite{kariyappa2019improving} & 91.1\% & 32.2\% &7.1\% &0.5\% &0.1\% &0.1\% &0\% &0\% \\
    GAL/8 \cite{kariyappa2019improving} & 92.4\% & 38.8\% &9.4\% &0.9\% &0.2\% &0\% &0\% &0\% \\ 
    \midrule
    \ours{}/3 & 91.3\% & 37.4\% &10.2\% &2.2\% &0.4\% &0.2\% &0\% &0\% \\
    \ours{}/5 & 91.9\% & 47.7\% &19.0\% &5.2\% &0.7\% &0.2\% &0.1\% &0\% \\
    \ours{}/8 & 91.1\% & 56.5\% &26.3\% &10.7\% &2.8\% &0.5\% &0.2\% &0.1\% \\
    \midrule
    \midrule
    AdvT/3 \cite{madry2018towards} &77.2\% &69.1\% &59.2\% &48.2\% &36.1\% &26.1\%	&17.2\%	&9.5\% \\
    AdvT/5 \cite{madry2018towards} &78.6\% &69.6\% &59.5\% &48.5\% &36.5\% &26.2\%	&16.0\%	&9.2\% \\
    AdvT/8 \cite{madry2018towards} &79.4\% &70.9\% &60.8\% &48.9\% &37.0\% &26.6\%	&17.1\%	&9.6\% \\
    \midrule
    \ours{}+AdvT/3 & 81.4\% & 71.4\% &59.1\% &44.1\% &30.4\% &19.8\% &11.1\% &5.5\% \\
    \ours{}+AdvT/5 & 83.4\% & 73.2\% &59.2\% &42.2\% &28.0\% &17.2\% &8.3\% &3.6\% \\
    \ours{}+AdvT/8 & 85.0\% & 72.3\% &57.8\% &40.8\% &25.7\% &14.8\% &6.7\% &3.1\% \\
    \bottomrule     
    \end{tabular}
\end{table}

\subsection{Numerical results for robustness}
\label{ap:rob_number}

We report numerical results that correspond to Figure \ref{fig:rob} and Figure \ref{fig:rob_ours_adv} in Table \ref{tab:ap_bbox_result} (black-box transfer accuracy) and Table \ref{tab:ap_wbox_result} (white-box accuracy), respectively. 
Note that ADP presents higher white-box accuracy than black-box transfer accuracy in some cases, e.g., 4.9\% > 0.4\% for ADP/8 when $\epsilon$ is 0.02, which implies ADP might result in obfuscated gradients \cite{athalye2018obfuscated}.

\begin{figure}[!htb]
\centering
\captionsetup{width=0.9\linewidth}
  \includegraphics[width=\linewidth]{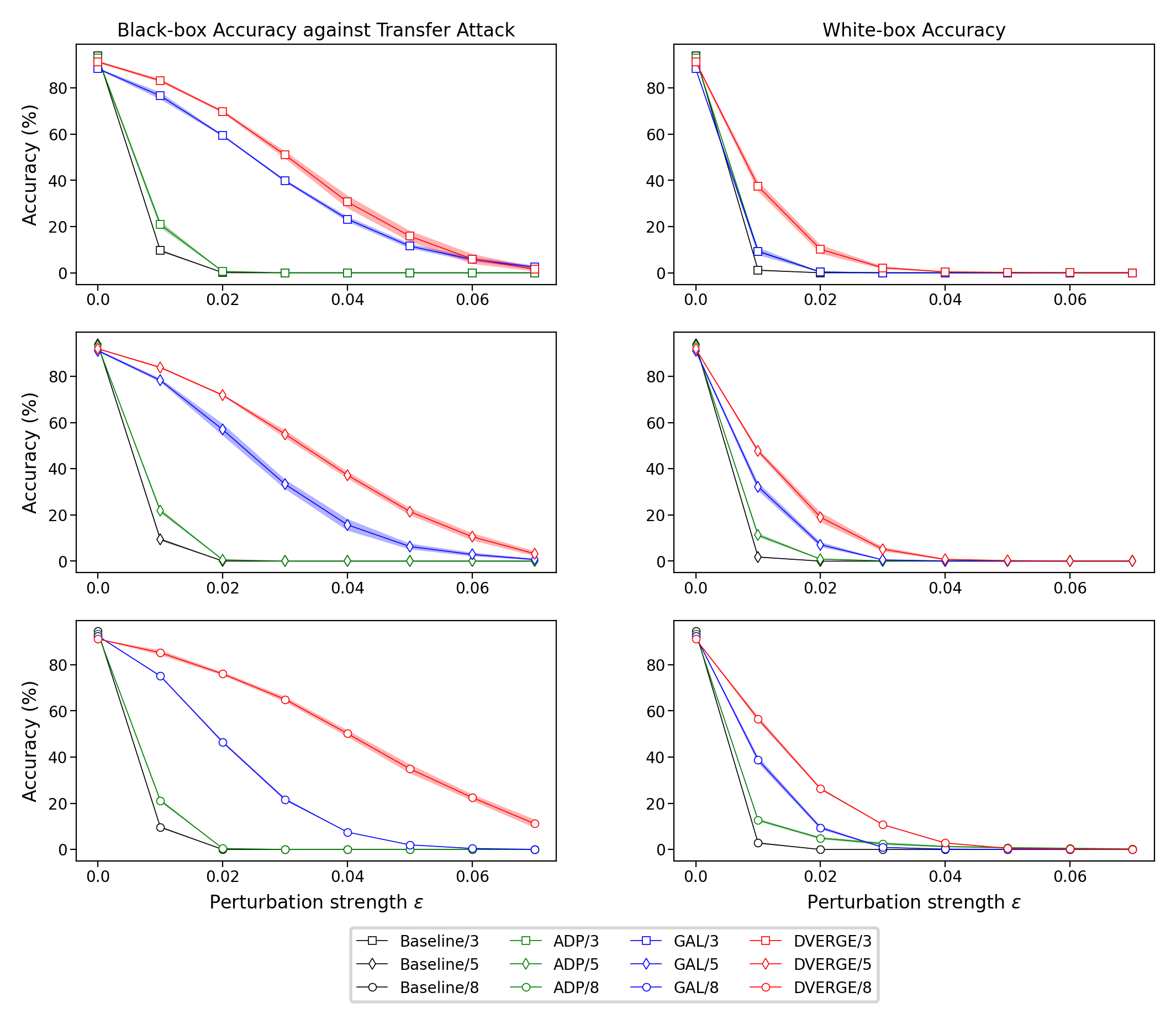}
  \caption{Robustness results with error bars for different ensemble methods. The number after the slash stands for the number of sub-models. From top to bottom, sub-plots on each row report the performance with 3, 5, and 8 sub-models respectively.}
  \label{fig:ap_rob_with_error_bar}
\end{figure}

In addition, we provide the robustness plots with error bars (indicating standard deviation) computed over three independent runs in Figure \ref{fig:ap_rob_with_error_bar}.
\ours{} presents a little higher variation in results, which we suspect is due to the random distillation layer selected in the last training epoch.
Refer to \textbf{Appendix \ref{ap:layer_selection}} for discussion on the layer effects.
However, \ours{} still yields noticeable improvements over other methods across the attack spectrum.

A more challenging scenario for adversarial defenses is assuming the attacker is fully aware of the exact defense that the system relies on.
In such case, the adversary can train an independent copy of the defended network as the surrogate model.
We report black-box transfer accuracy of each method under this setting in Table \ref{tab:ap_bbox_result_against_ours}.
The same group of attacks as in Section \ref{sec:4.3} are used and ensembles with 3, 5, and 8 sub-models form the collection of surrogate models.
According to Table \ref{tab:ap_bbox_result_against_ours}, \ours{} still presents the strongest robustness against such a powerful black-box transfer adversary among all ensemble methods.

\begin{table}[!hbt]
    %\small
    \footnotesize
    \caption{Accuracy v.s. $\epsilon$ against black-box transfer attacks generated from hold-out ensembles that are trained with the exact defense technique used by each ensemble. The number in the first column after the slash is the number of sub-models within the ensemble.}
    \label{tab:ap_bbox_result_against_ours}
    \centering
    \begin{tabular}{c|cccccccc}
    \toprule
    $\epsilon$ & clean & 0.01 & 0.02 & 0.03 & 0.04 & 0.05 & 0.06 & 0.07 \\
    \midrule
    \midrule
    ADP/3 \cite{pang2019improving} & 93.3\% & 34.7\% &6.5\% &1.4\% &0.2\% &0\% &0\% &0\% \\ 
    ADP/5 \cite{pang2019improving} & 93.1\% & 34.2\% &6.6\% &1.6\% &0.6\% &0\% &0\% &0\% \\ 
    ADP/8 \cite{pang2019improving} & 93.0\% & 32.5\% &5.6\% &1.4\% &0.2\% &0\% &0\% &0\% \\ 
    \midrule
    GAL/3 \cite{kariyappa2019improving}
    & 88.8\% & 67.8\% &36.2\% &13.8\% &3.7\% &0.6\% &0.1\% &0\% \\ 
    GAL/5 \cite{kariyappa2019improving} & 91.0\% & 67.5\% &31.9\% &9.2\% &1.8\% &0.3\% &0\% &0\% \\
    GAL/8 \cite{kariyappa2019improving} & 92.3\% & 64.7\% &25.8\% &5.4\% &0.6\% &0.1\% &0\% &0\% \\ 
    \midrule
    \ours{}/3 & 91.4\% & 75.4\% &50.2\% &23.8\% &7.6\% &2.1\% &0.3\% &0.2\% \\
    \ours{}/5 & 91.9\% & 77.2\% &53.1\% &26.7\% &9.5\% &2.6\% &0.5\% &0.2\% \\
    \ours{}/8 & 91.1\% & 77.7\% &57.3\% &32.0\% &13.9\% &3.8\% &0.9\% &0.2\% \\
    %\midrule
    %\ours{}+AdvT/3 & 83.0\% & 78.1\% &74.2\% &68.1\% &60.7\% &51.5\% &41.5\% &33.3\% \\
    %\ours{}+AdvT/5 & 84.5\% & 80.2\% &74.4\% &65.3\% &55.5\% &43.9\% &33.5\% &23.7\% \\
    %\ours{}+AdvT/8 & 85.9\% & 78.6\% &71.1\% &61.7\% &50.4\% &39.2\% &28.6\% &18.7\% \\
    \bottomrule     
    \end{tabular}
\end{table}

\subsection{Discussion on \ours{} with adversarial training}
\label{ap:ours_plus_adv}

\begin{figure}[!htb]
\centering
  \includegraphics[width=\linewidth]{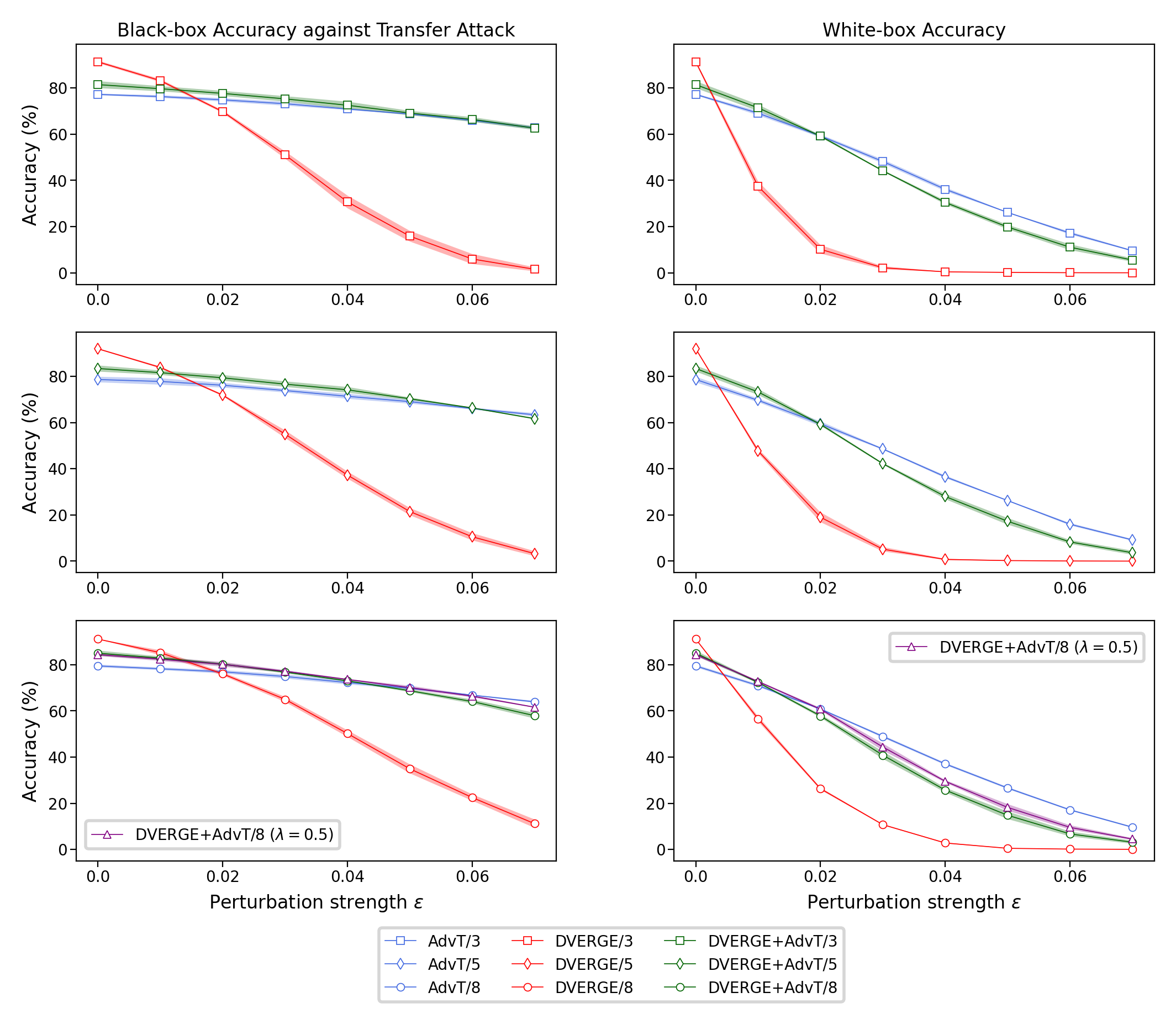}
  \captionsetup{width=0.9\linewidth}
  \caption{Results for \ours{} combined with adversarial training. From top to bottom, sub-plots on each row report the performance with 3, 5, and 8 sub-models respectively.}
  % Need to add how the axis are chosen
  \label{fig:ap_plus_adv}
\end{figure}

Formally, the combined training objective of \ours{} and adversarial training is

\begin{equation}
    \label{equ:plus_adv}
    \min_{f_i} \mathbbm{E}_{(x,y),(x_s,y_s),l} \left[\underbrace{\lambda\cdot\sum_{j\neq i}\mathcal{L}_{f_i}(x'_{f_j^l}(x,x_s),y_s)}_{\text{\ours{} loss}}+\underbrace{\vphantom{\lambda\cdot\sum_{j\neq i}\mathcal{L}_{f_i}(x'_{f_j^l}(x,x_s),y_s)} \max_{\delta\in\mathcal{S}}\mathcal{L}_{f_i}(x_s+\delta,y_s)}_{\text{AdvT loss}}\right],
\end{equation}
where $\lambda$ is a hyperparameter that balances between the two terms.
We set $\lambda=1.0$ to achieve the results in Figure~\ref{fig:rob_ours_adv}.
Results for other ensemble sizes under $\lambda=1.0$ are shown in Figure \ref{fig:ap_plus_adv}, where the standard deviation over three independent runs is also included.
The observations stay the same as in Section \ref{sec:4.4}.
To better reflect the trade-off between clean accuracy and robustness, we also report the results for an ensemble of eight sub-models with $\lambda=0.5$.
In this case, \ours{} loss is weighted less and the training process will favor adversarial training.
In turn, the ensemble spends more of its capacity to capture robust features instead of diverse non-robust features. 
Consequently, compared with $\lambda=1.0$, we observe a decrease in clean accuracy and an increase in both black-box and white-box robustness when $\epsilon$ is large.
In addition, the ensemble size is actually another weight factor in Equation (\ref{equ:plus_adv}) as increasing the number of sub-models will naturally lead to larger \ours{} loss such that it outweighs the AdvT loss.
As a result, larger (smaller) ensemble sizes for \ours{}+AdvT results in better (worse) clean performance yet worse (better) black-box and white-box robustness under a large $\epsilon$. 
This assertion can be confirmed by the results in the bottom three rows of Table \ref{tab:ap_bbox_result} and Table~\ref{tab:ap_wbox_result}.

\subsection{Ablation study on layer selection for distillation}
\label{ap:layer_selection}
As aforementioned, at each epoch of the DVERGE training, we randomly sample a layer from 20 candidate layers in the ResNet-20s to perform feature distillation based on the intuition that this could help avoid overfitting to any specific layer's feature sets. 
However, it is natural to wonder what the difference is between using each individual layer.
While a rigorous investigation on layers’ effect will be an important future direction for this work, here we compare the random layer selection with a straightforward alternative which is using a fixed layer for distillation throughout the training.

\begin{table}[!htbp]
\footnotesize
\centering
\caption{Black-box robustness ($\epsilon$=0.03) / white-box robustness ($\epsilon$=0.01) of each layer choice. The results for random layer selection are directly taken from Table \ref{tab:ap_bbox_result} and \ref{tab:ap_wbox_result}.}
\begin{tabular}{ccccc}
\toprule
ResBlock 1 &ResBlock 2 &ResBlock 3 &Output Layer &Random\\
\midrule
50.7\%~/~39.3\% 
&50.2\%~/~36.5\%
&32.8\%~/~31.0\%
&37.3\%~/~33.5\%
&51.0\%~/~37.4\%\\
\bottomrule
\end{tabular}
\label{tab:ap_layer_choice}
\end{table}

\begin{figure}[!htb]
\centering
\captionsetup{width=0.9\linewidth}
  \includegraphics[width=.9\linewidth]{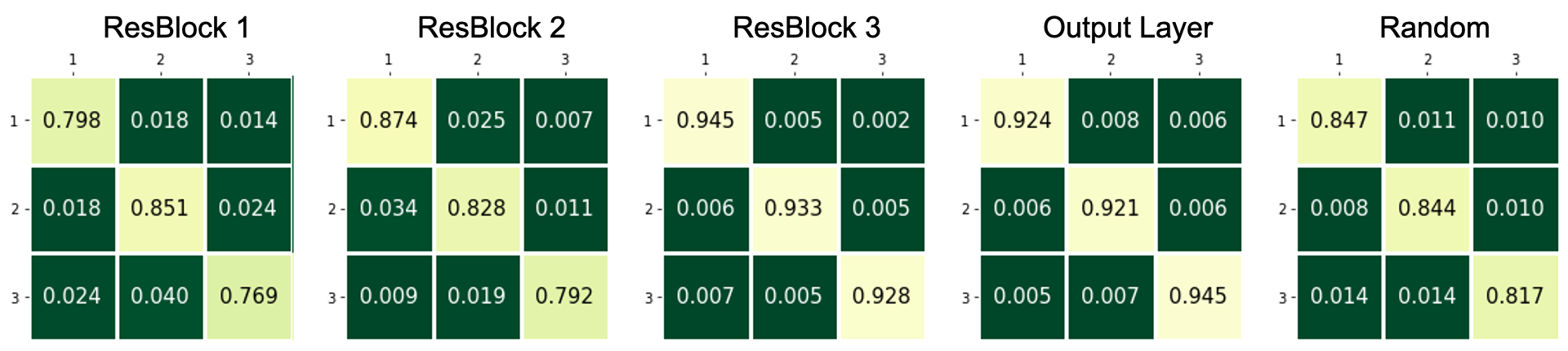}
  \caption{Pair-wise transferability results for different layer selection.}
  \label{fig:ap_layer_choice}
\end{figure}

Specifically, we train another four ensembles by distilling only from the output layer of either ResBlock 1, ResBlock 2, ResBlock 3, or the whole network (correspondingly, the 7-th, 13-th, 19-th, and 20-th layer of the ResNet20). 
Then we report the black-box and white-box robustness results in Table~\ref{tab:ap_layer_choice} and present the pair-wise transferability results (evaluated under $\epsilon$=0.01) in Figure \ref{fig:ap_layer_choice}.
It can be seen that while training with deep layers (ResBlock 3 and the output layer) could indeed lead to lower transferability between sub-models, training with shallow layers (ResBlock 1 and 2) can introduce a noticeably higher white-box robustness for each sub-model itself, as shown by the lower diagonal numbers in the first and second heatmap of Figure~\ref{fig:ap_layer_choice}. 
The fact that some degree of white-box robustness can be achieved with shallow layers is surprising and needs further study, but this observation explains why they lead to stronger overall robustness for the ensemble than deep layers in Table~\ref{tab:ap_layer_choice}. 
Meanwhile, training with random layers can balance the effect from both shallow and deep layer, achieving both individual robustness improvement and low transferability at the same time. 
These two factors both contribute to the overall black-box robustness of the ensemble. As shown in Table \ref{tab:ap_layer_choice}, random layer selection provides superior black-box robustness than other choices, while the achieved white-box robustness is a little lower than the best alternative (ResBlock 1) due to the lack of individual robustness in each sub-model.

\subsection{Ablation study on pre-training}
\label{ap:pre-training}
\begin{figure}[!htb]
\centering
\captionsetup{width=0.9\linewidth}
  \includegraphics[width=\linewidth]{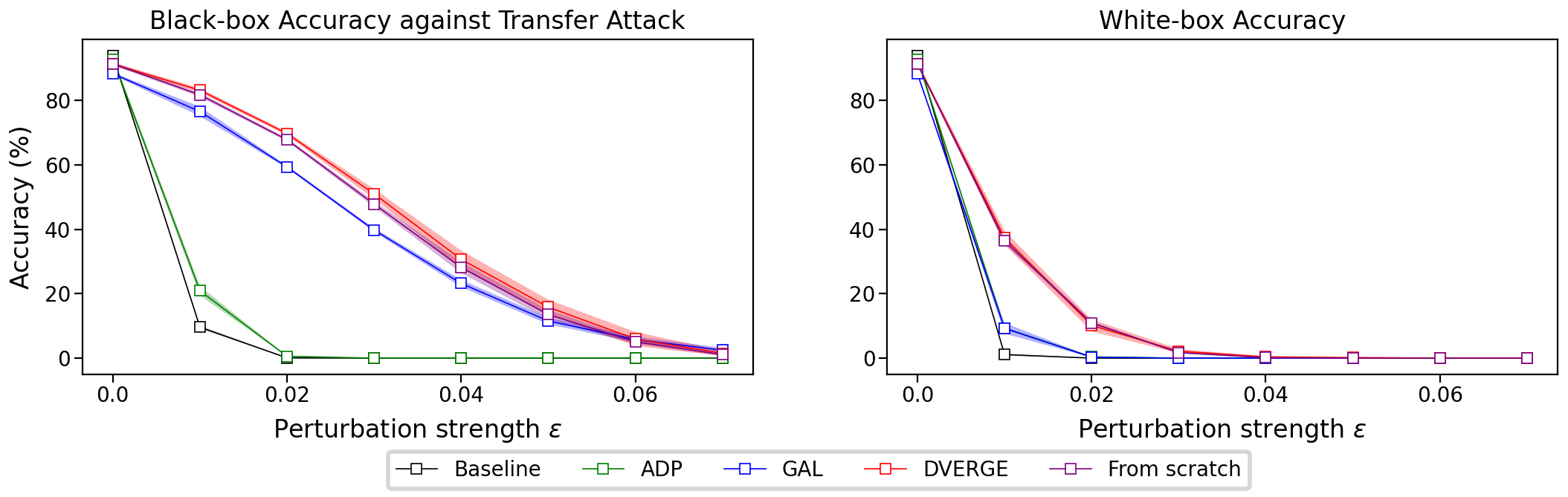}
  \caption{Results of \ours{} with or without pre-training comparing to other methods. All results are evaluated with 3 sub-models. Error bars are evaluated with 3 repeated runs.}
  \label{fig:ap_pretrain}
\end{figure}

As mentioned in Section \ref{ssec:alg}, we train \ours{} from a pre-trained ensemble based on the intuition that well-learnt features of the pre-trained models are more informative for distillation and diversification.
However, we show in Figure~\ref{fig:ap_pretrain} that although slightly worse than using pre-trained models, training from scratch still offers improved robustness over others, implying that pre-training is not strictly necessary for \ours{}.

\section{Convergence check}
\label{ap:convergence}

As suggested in \cite{tramer2020adaptive,carlini2019evaluating}, we report accuracy vs. the number of attack iterations in Table \ref{tab:ap_convergence_check}. 
Note, we use only one random start here for white-box attacks for efficiency.
One can observe that using more steps decreases the accuracy by no more than 0.6\% for black-box attacks and no more than 1.2\% for white-box attacks. Thus, we confirm sufficient steps have been applied and all attacks have converged during the evaluation.

\begin{table}[htb]
    %\small
    \footnotesize
    \caption{Accuracy against attacks with varying number of iterations.}
    \label{tab:ap_convergence_check}
    \centering
    \begin{tabular}{c|ccc|ccc}
    \toprule
     & \multicolumn{3}{c}{black-box ($\epsilon=0.03$)} & \multicolumn{3}{c}{white-box ($\epsilon=0.01$)} \\
    \cmidrule(lr{0.5em}){2-4} \cmidrule(lr{0.5em}){5-7}
     & 100 & 500 & 1000 & 50 & 500 & 1000 \\\midrule
    \ours{}/3 &53.2\% &52.6\% &53.4\% &42.7\%~ &41.6\% &41.5\% \\
    \ours{}/5 &57.2\% &56.9\% &57.3\% &50.4\% &49.5\% &49.5\%\\
    \ours{}/8 &63.6\% &63.6\% &63.2\% &58.0\% &57.8\% &57.8\%\\
    \ours{}+AdvT/3 &76.2\% &76.3\% &76.2\% &72.9\% &72.9\% &72.9\%\\
    \ours{}+AdvT/5 &77.9\% &78.0\% &77.8\% &74.8\% &74.8\% &74.8\%\\
    \ours{}+AdvT/8 &77.1\% &76.9\% &77.4\% &72.7\% &72.7\% &72.7\%\\
    \bottomrule     
    \end{tabular}
\end{table}

\end{document}